\documentclass[article,3p,authoryear,times]{elsarticle}
\usepackage{graphicx}
\usepackage{subcaption}
\usepackage[fleqn]{amsmath}
\usepackage[]{amsfonts}
\usepackage[]{amsthm}
\usepackage[]{amssymb}
\usepackage{empheq}
\usepackage{rotating,tabularx}
\usepackage{multirow}
\usepackage{geometry}
\usepackage[export]{adjustbox}
\usepackage{wrapfig}
\usepackage{siunitx}
\usepackage{xcolor}
\usepackage{multicol}
\usepackage{siunitx}

\usepackage{caption}
\usepackage{graphicx, stmaryrd, url}

\usepackage{bm}
\usepackage{glossaries}
\usepackage{amsmath}
\usepackage{float}

\makeglossaries
\newglossaryentry{lengthtree}%
{%
  name={$L_t$},
  text={L_t},
  description={description here}
  sort={L}
}

\usepackage{ifthen}
\newboolean{finalversion}
\setboolean{finalversion}{true}
\setboolean{finalversion}{false}

\usepackage{soulutf8}
\usepackage{color}
\usepackage{xargs}
\newcounter{commentnumber}
\newcommandx{\comment}[3][1=,2=]{%
  \hl{$^{\arabic{commentnumber}}$}\stepcounter{commentnumber}%
  \hl{#1}%
  \ifthenelse{\equal{#2}{}}%
  {}%
  {\hl{$\rightarrow$#2}}%
  \ifthenelse{\equal{#1}{}\AND\equal{#2}{}}%
  {}%
  {\hl{: }}
  \hl{#3}}
\iffinalversion
\renewcommandx{\comment}[3][1=,2=]{}
\fi

\usepackage[mathlines]{lineno}
\usepackage[english]{babel}
\usepackage[T1]{fontenc}

\usepackage{hyperref} % To include references inside the pdf file
\usepackage{graphicx}
\usepackage{tikz}
\usepackage{pstricks}
\usepackage{xstring}
\usepackage{multirow}

\definecolor{darkblue}{rgb}{0,0,0.5}

\hypersetup{colorlinks=true,linkcolor=black,citecolor=darkblue,urlcolor=darkblue} % Set options for the hyperref package

\newcommand{\newref}[2]{\hyperref[#2]{#1~\ref*{#2}}} % hyperref: \newref{labeltext (e.g. table, figure, ...)}{labelname}

\begin{document}

\begin{frontmatter}

\title{A methodological framework for Resilience as a Service (RaaS) in multimodal urban transportation networks}

\author[GRETTIA,VEDECOM]{Sara Jaber\corref{cor}}
\author[GRETTIA]{Mostafa Ameli}
\author[VEDECOM]{S. M. Hassan Mahdavi}
\author[GRETTIA]{Neila Bhouri}

\address[GRETTIA]{Univ. Gustave Eiffel, COSYS, GRETTIA, Paris, France}
\address[VEDECOM]{VEDECOM, mobiLAB, Department of new solutions of mobility services and shared energy, Versailles, France}

\begin{abstract}

Public transportation systems are experiencing an increase in commuter traffic. This increase underscores the need for resilience strategies to manage unexpected service disruptions, ensuring rapid and effective responses that minimize adverse effects on stakeholders and enhance the system’s ability to maintain essential functions and recover quickly. This study aims to explore the management of public transport disruptions through resilience as a service (RaaS) strategies, developing an optimization model to effectively allocate resources and minimize the cost for operators and passengers. The proposed model includes multiple transportation options, such as buses, taxis, and automated vans, and evaluates them as bridging alternatives to rail-disrupted services based on factors such as their availability, capacity, speed, and proximity to the disrupted station. This ensures that the most suitable vehicles are deployed to maintain service continuity. Applied to a case study in the Ile de France region (Paris and its suburbs), complemented by a microscopic simulation, the model is compared to existing solutions such as bus bridging and reserve fleets. The results highlight the model’s performance in minimizing costs and enhancing stakeholder satisfaction, optimizing transport management during disruptions. 

\end{abstract}


\end{frontmatter}

%%%%%%%%%%%%%%%%%%%%%%%%%%%%%%%%%%%%%%%%%%%%%%%%%%%%%%%%%%%%%
%% CONTENTS
%%%%%%%%%%%%%%%%%%%%%%%%%%%%%%%%%%%%%%%%%%%%%%%%%%%%%%%%%%%%%

% \linenumbers % Running line numbers.
%\pagewiselinenumbers % comment out for final manuscript

%%%%%%%%%%%%%%%%%%%%%%%%%%%%%%%%%%%%%%%%%%%%%%%%%%%%%%%%%%%%%

\section{Introduction}

Public transport systems are integral to urban life and provide essential mobility services to millions of people. In France, nearly 70\% of Parisians rely on the metro, buses, or suburban trains for their daily commutes. \cite{stat2020} underscored the critical role of these systems in facilitating urban connectivity, particularly in densely populated regions like Île de France, including Paris and suburban areas.
However, these systems are inherently vulnerable to disruptive events ranging from technical failures to natural disasters \citep{zhang2023quantitative}. Such vulnerabilities have been extensively analyzed in recent studies focusing on the resilience and vulnerability of urban rail transit systems under various disruptive scenarios \citep{huang2023vulnerability, xu2024resilience}. The cost of disruptions extends beyond operational delays, which encompass the economic losses and reputational damage incurred by operators due to service interruptions, the time cost to passengers resulting from delayed journeys, and the potential strain on public authorities managing the aftermath \citep{de2016transit}. 
According to \cite{jenelius2020resilience}, disruptions within the public transport system can occur unexpectedly, potentially triggering cascading effects on both the demand and supply sides. For instance, \cite{yap2019analysis} observed that the Washington DC metro network experiences an average of 20 incidents daily, leading to train or line delays of a minimum of two minutes. Similarly, studies have highlighted the resilience challenges in global metro systems during unexpected events, emphasizing the need for comprehensive disruption management frameworks \citep{zhang2023quantitative, wang2024modelling}. Furthermore, it is imperative for public transport providers to not only devise proactive strategies for such disruptions but also respond promptly to these occurrences, ensuring the continuity of a resilient service. Resilience refers to the ability of a public transport system to effectively minimize both the scale and duration of disruptions and return to its normal performance \citep{chopra2016network}. This definition aligns with recent research emphasizing the critical role of resilience in maintaining service continuity and mitigating the impacts of disruptions on urban transport systems \citep{sajjad2021rethinking}. In this study, we applied our disruption-management model to the rail system, which is a critical component of urban public transport networks and often faces significant challenges during service interruptions. We aim to address disruptions proactively and use them as opportunities to reinforce the resilience of the transport system. To achieve this goal, dispatching policies for alternative transport modes must be established, which can be challenging because various factors must be considered, including the nature, location, and duration of the disruption, as well as the resources available to the mobility service provider. Several strategies can be employed to provide alternatives for passengers, such as deploying a reserve fleet of buses for bridging services, which may appear to be a viable interim measure during disruptions. Using this method leads to time delays and passenger dissatisfaction, as noted in \cite{ zeng2012collaboration}. Another potential solution is to deploy taxis to provide a more flexible and faster replacement strategy for disrupted transport services, which could reduce the waiting duration of passengers and avoid delays. However, taxis may be more expensive than buses, and deploying numerous taxis to provide replacement services could be prohibitive, as discussed in \cite{cebecauer2021integrating}. These solutions may require significant financial investment for resource allocation, maintenance, and planning for a reserve fleet or considerable arrangement costs for using external fleets with decentralized management. Reflecting on the essential role of resilience in urban transportation, \cite{mahdavi2024synthetic, luo2021towards} emphasized its importance in overcoming operational challenges and ensuring effective service continuity. 

To address these challenges, the concept of resilience as a service (RaaS) introduced by \cite{ amghar2023resilience} proposed a collaborative mechanism among various mobility service providers to minimize the cost of creating replacement services during disruptions. Unlike traditional bus bridging, which relies on reserve buses from depots, RaaS utilizes existing resources such as in-service vehicles. This integrated framework leverages existing services without requiring additional resources. 
This study aimed to develop a mathematical model for RaaS that optimizes resource reallocation while minimizing disruption costs and reducing passenger inconvenience. In \cite{amghar2023resilience}, the concept of RaaS is defined and proposed as a preliminary solution. In this study, we provide a comprehensive framework for multimodal urban transportation, introducing a novel strategy based on a cost-benefit analysis that responds to disruptions using available resources while considering operational costs. The proposed model was designed to evaluate alternative solutions when a particular mobility mode encounters disruptions, to ensure that the most affordable and efficient options are pursued. Collaboration with diverse transport entities, including buses, taxis, and automated vans, is required to offer replacement services that cover all affected stations with adequate capacity. The model aims to determine the optimal number of vehicles required to restore normal service while ensuring that reallocating public transport vehicles, such as buses, do not use lines with long headways, thus preventing significant disruptions at the stations of collaborating transport lines. In addition, we addressed the complexities of urban multimodal networks by employing a simulation-based approach to account for congestion. Using agent-based simulation models significantly enhances our understanding of passenger behavior during unforeseen events, a level of detail that is often unattainable through purely analytical methods \citep{leng2018agent}. \\
The remainder of this paper is organized as follows: Section 2 provides a literature review on public transport disruption management strategies.  Section 3 presents the RaaS methodological approach for optimizing service resilience through the integration of alternative vehicles. Section 4 elaborates on the problem statement, delineating the RaaS solution, stakeholders involved, and overarching objectives. Section 5 presents our optimization model, which encompasses network and disruption modeling, as well as a cost analysis to establish the replacement process for a disruptive event. Section 6 describes the case study and assumptions and outlines the specific scenarios, parameters, and assumptions utilized in our simulation and optimization processes. In Section 7, we analyze and discuss the results obtained, supplemented by a sensitivity analysis, and offer insights into the efficacy of the RaaS framework. Finally, Section 8 concludes the paper by summarizing our contributions and discussing future implications.

\section{Literature review}

In this section, we provide a structured exploration of previous studies on disruption management in public transportation systems. By organizing the studies into distinct subsections, those that explored conventional and bridging approaches to disruption management highlighted the progress of research efforts from foundational approaches to innovative solutions.

\subsection{Conventional approach: adjusting rolling stock and timetables}

Research addressing disruptions in public transport systems often concentrates on rail services, with conventional strategies typically involving adjustments to the rolling stock and timetables. For example, \cite{cadarso2013recovery} introduced a heuristic mixed-integer linear programming (MILP) model for optimizing train schedules and rolling stock in response to urban rail network disruptions, with a focus on passengers' reactions to the disruption. Recovery strategies involve either the cancellation of current train services or the addition of new ones. However, the option of rescheduling the timing of existing trains was not considered. Besides, \cite{veelenturf2017passenger} employed a heuristic iterative framework that acknowledges passenger demand for the rescheduling of both rolling stock and timetables into disruption management. Their approach modified timetables to incorporate additional stops. However, only train stops at stations that they normally would not call can be used in timetable decisions, and passengers need to adapt their paths to the new schedule.  \cite{zhan2015real} proposed a two-stage model for rescheduling train timetables, aimed at minimizing the total delay caused by a complete railway blockage. This model integrates stop planning as a key constraint and is structured as a MILP model to prevent the cancellation or rerouting of en-route trains. This study specifically examined the scheduling challenges posed by a complete blockage of railways for high-speed trains, which may not encompass partial blockages or other types of rail services. Further research has expanded on these models, incorporating resilience assessments to address the broader impacts of such disruptions on urban transportation systems \citep{chen2023vulnerability}. \cite{kroon2015rescheduling} tackled the challenge of large-scale disruptions and presented an approach to real-time rolling stock rescheduling amidst railway disruptions, offering a simulation model integrated with passenger flow dynamics to optimize the schedules and minimize passenger delays. The authors developed a two-step simulation methodology, initially simulating passenger traffic dynamics and subsequently integrating the results into a rolling-stock rescheduling model. Although this minimizes passenger delays, this study did not analyze the related operational costs of rescheduling. Additionally, the model demonstrated heightened effectiveness during peak hours with a constrained seat capacity; however, its efficacy may diminish during off-peak periods. \cite{zhan2016rolling} concentrated on the rescheduling of double-track high-speed trains in case of partial blockage, with the duration of the blockage being uncertain. They developed a MILP model and employed a rolling-horizon algorithm to address the issue. Although this model can yield an optimized timetable, it has computational complexity and slow response, particularly for extensive railway networks. \cite{ghaemi2018macroscopic} extends the rescheduling of mixed integer programming (MIP) models from macroscopic to a microscopic level during complete blockages. When applied to real test cases, the model can include timetables and transition plans by employing short-turning, partial cancellation, and rescheduling strategies and considering the impact of disruption duration. Owing to the complexity of the problem, this study focused primarily on addressing a single disruption at a time. Based on their previous work \cite{zhan2016rolling}, \cite{zhan2021integrated} integrated a train rescheduling and passenger routing problem while considering seat capacity and train routing at the microscopic level using the alternating direction method of multipliers for decomposition solved with a dynamic programming approach, to minimize the cost of each passenger on travel duration. However, disrupted trains must wait until the disruption ends, which leads to long delays for passengers. \cite{shakibayifar2020integrated} focused on minimizing train delays and deviations from the set timetable, assigning weights to these delays based on passenger demand at each station. They utilized the MIP model to represent railway infrastructure, assuming a limited station capacity. To address this problem, they proposed an exact solution method and two heuristic approaches: right-shift and two-stage rescheduling. Nevertheless, this solution was effective only for short-duration track blockages, as it could not provide a solution within a reasonable timeframe for longer disruption durations. \cite{hassannayebi2021simulation} introduced a decision support system employing a simulation-optimization framework aimed at reducing the average wait duration for passengers by implementing rescheduling strategies such as short-turn and skip-stop methods. Their approach was supplemented with a variable neighborhood search algorithm with a non-stationary Poisson arrival rate of passengers. However, their solution did not capture the inter-decision dynamics between mobility service operators and commuters. 
However, adjusting rolling-back and timetable strategies is ineffective for promptly addressing the direct impacts of service disruptions, resulting in significant system-wide delays. This is because of the time gap between decision-making by operators and the execution of recovery actions, as discussed in \cite{kepaptsoglou2009bus}. Additionally, these studies address disruption management only from a passenger-centric perspective. To overcome these challenges, a bridging approach has emerged to mitigate the impacts of unplanned disruptions.

\subsection{Bridging strategy: bus, taxi, and multimodal}

This subsection explores the bridging strategy, which restorers connectivity between disrupted segments of transportation networks, using alternative transport means such as buses, taxis, and multimodal services to maintain service continuity \citep{wang2023integrated}. This approach has been increasingly recognized in the literature, with studies highlighting the benefits of multimodal integration and the challenges of implementing such strategies effectively in dense urban environments \citep{chen2024intelligent}. \cite{kepaptsoglou2009bus} advocate for the use of bus bridging service for rail disruption management. Their contributions included a three-tiered modeling framework aimed at defining the bus bridging environment, designing bus routes, and allocating resources. They emphasized passenger welfare by considering demand patterns and resource constraints when designing bus-bridging routes. However, their approach did not account for the complexities of long-term or system-wide disruptions. \cite{jin2016optimizing} utilized a column generation algorithm to generate potential bridging routes to minimize the total increase in travel duration of all commuters in disruption compared to normal situations. They employed an integrated optimization model to perform route selection, frequency determination, and bus allocation. This study applies a demand-responsive approach if the bus routes originate from a bus node located at the end of the disrupted segment. Although the column-generation algorithm effectively generates demand-responsive candidate bus routes, its scalability may be limited when applied to disruptions that affect extensive network segments. Additionally, the model's focus on optimizing travel time reductions might inadvertently neglect certain bus routes that, although not optimal in terms of travel time, could be critical for servicing remote or less connected areas. This oversight can affect the comprehensive coverage of the network, potentially leaving some areas underserved during critical disruption events. 

Rather than designing route choice models, \cite{codina2013model} developed a mathematical nonlinear integer programming (NIP) model and a heuristic solution method for planning a bus-bridging system. The model considered deciding bus allocation on each bus bridging route and commuter traveling route under congestion, to minimize travel duration. using automatic fare collection data, they assumed the existence of a set of predefined bus bridging routes. This study primarily addressed the micro-level operational aspects of bus bridging systems, including commuter queuing flows at bus stops and the available space for bus queuing. However, their approach focused only on existing bridging routes, which did not dynamically adapt to unforeseen disruption events. In contrast to approaches that rely on static bridging systems, \cite{gu2018plan} developed a two-stage NIP model to create a flexible bus bridging plan that allows for the dynamic sequencing of stations visited by each bus. This model aimed to balance bus operations with commuter needs and incorporate a rolling-horizon framework to manage dynamic passenger arrivals during rail disruptions. This approach enables buses to serve varying routes and optimizes both bridging time and passenger delay. In practice, the proposed model may result in queuing buses owing to limited spaces at the affected stations and increase the travel duration when numerous bridging buses run on the same link. \cite{de2012evaluating} measured rail network robustness from the user’s perspective, comparing system behaviors under disruptions with and without bus bridging services. This study measured passenger travel duration and flow before and after rail failures to evaluate and assess how bus bridging reduces the total travel duration of the network. However, this study does not explore broader aspects of network performance, such as overall operational efficiency or the financial implications of disruptions. \cite{wang2014bus} examined the impact of random commuter demand on bus bridging scenarios, employing compound Poisson processes including balking and reneging, and conducting large-scale Monte Carlo simulations to explore passenger behavior in closed public transport systems under budget constraints. However, their study primarily focused on demand modeling and did not address the bus-bridging network design, which is a central aspect of bus-bridging services. Another study conducted under budget constraints by \cite{van2016shuttle} developed a model to optimize shuttle bus lines and frequencies, aiming to minimize commuter inconvenience costs, including transfer- and frequency-dependent wait duration costs. An MIP formulation based on a path-based multicommodity flow model was used to determine the optimal set of bridging bus routes and their frequencies. However, this model assumes that demand and frequency are not time-dependent. Although this assumption is common in high-frequency networks with cyclic timetables, it may not be valid for all public transportation networks.  

Besides, \cite{yang2017optimizing} investigated the bus bridging problem for rail transit system congestion, proposing a two-stage mathematical modeling procedure. They developed a two-objective timetable model and designed a binary coding genetic algorithm to solve the IP model to minimize the passenger wait duration. This study did not incorporate the concept of time intervals to describe the evolution of boarding/alighting passengers from a network-level perspective. \cite{zhang2018metro} addressed the determination of substitute bus service initiation time under uncertain railway disruption recovery time, aiming to minimize the combined costs of bus initiation time and commuter delay. They developed an optimization model to evaluate the optimal initiation time of a substitute bus service, incorporating demand considerations and modeling the disruption duration as a probability distribution. By focusing on metro disruption management, this study highlights the need to integrate alternative transportation modes for rapid system recovery, particularly for rail systems with overwhelming commuter demand. The research did not evaluate alternative operational plans within or across different modes of public transport, limiting its scope to the optimization of bus-bridging services under uncertainty. \cite{zhang2020metro} investigated various service contract designs between metro and bus operators under uncertain recovery time of unplanned metro service disruption, focusing on the benefits and costs to both parties and highlighting the need for authorities’ intervention to regulate the pricing of substitute bus services to protect passenger welfare. The study employed the MILP model and graph-partitioning techniques to optimize the service area and route planning while considering two payment schemes for the bus company, fixed payment and linear payment, with a quantification of the willingness of passengers to wait for replacement vehicles. While this study provides a framework and findings that can guide the operators toward contractual passenger-oriented arrangements, it does not explicitly state a single compromise solution. \cite{pender2014improving} raise a pertinent issue regarding the location of spare buses, which are utilized for bridging services during disruptions. Although the common practice is to station these vehicles at existing depots, the authors suggest that alternative locations may also be considered for this purpose. By introducing the concept of satellite or virtual bus depots for rapid deployment during unplanned rail disruptions, this study proposes optimizing bus reserve depot locations by minimizing the total demand-weighted travel duration between the demand and facility, assuming a fixed bus fleet size and capacity. The optimization of reserve fleet locations has been explored; however, the cost-effectiveness and challenge of driver availability of such a fleet remain unaddressed. 

\cite{pender2013disruption} examined instances of urban rail transit disruptions across many international transport organizations and found that the implementation of bus bridging services is the predominant strategy employed to address such interruptions. Although buses have been extensively studied for bridging purposes, the inclusion of other transportation modes, such as taxis, remains relatively understudied. Moreover, there is a significant lack of literature on multimodal network models that integrate various modes of transportation. 
\cite{zeng2012collaboration} investigated the use of taxis as an alternative recovery service of bus bridging strategy during tram system disruptions. Their study contributes to the development of decision functions for tram and taxi service providers to facilitate collaboration and establish service payment under fixed and linear schemes. They analyzed the reduced loss of the tram operator and the profit of the taxi operator to determine the feasibility of implementing a replacement process. However, their approach is limited by its applicability to disruptions with relatively small passenger volumes and shorter durations (e.g., less than 1h), requiring alternative bus services for longer-term disruptions. \cite{fang2015long} advanced the study of contracts for disruption recovery services between public transport and taxi operators. Their research employed a game-theoretic model with mathematical equations to ascertain the optimal number of taxis to reserve, examining the decision-making processes from the perspectives of both taxi and tram operators. However, this study did not explore how to balance the tram company’s dual objectives of enhancing passenger satisfaction and reducing recovery costs, nor did it investigate the possibility of integrating buses with taxis for a more comprehensive recovery service approach. \cite{cebecauer2021integrating} examined the combined use of bus bridging and on-demand taxi services to manage rail disruptions, focusing on optimizing the balance between service operators’ costs and passenger wait duration within the disrupted area. Their methodology involved a collaborative simulation model between taxis and bus bridging, focusing only on the disrupted location, and was partially validated in terms of passenger arrivals, in-vehicle times, and trip lengths. The study did not incorporate a model to assess the effects of the disruption strategy on traffic and service levels in the rest of the network or a business model for cooperation between service operators to establish the replacement process. \cite{fang2019replacement} has explored similar integration challenges, focusing on the dynamic allocation of multimodal resources and the environmental impacts of replacement services. \cite{yuan2018dynamic} introduced a dynamic framework designed to integrate public transportation resources like buses and taxis during disruptions. Their methodology involved the implementation of an ILP model that focused on the spatial rerouting and temporal rescheduling of multimodal transport vehicles. Using real-time automatic fare collection data, they optimized vehicle allocation and routing to maximize passenger transport capacity while minimizing rerouting and reallocation costs. However, passenger behaviors such as travel duration tolerance and mode-switching willingness remain unexplored. From an environmental perspective, \cite{fang2019replacement} investigated the integration of taxis as an alternative mode of transportation for tram and light-rail systems during disruptions. Their research involved a comparative analysis between traditional bus bridging and taxi replacement services, utilizing mathematical equations to examine the financial cost of taxi recovery services and the environmental loss calculated as additional carbon emissions when light rail service providers take no action to transfer stranded passengers while considering passenger behavior and disruption duration as significant factors. However, the study potentially oversimplified real-world scenarios and incompletely captured long-term economic and environmental impacts. \cite{fang2020disruption} focused on developing a decision support tool for tram companies to manage replacement services during short-term unplanned disruptions, considering critical factors such as commuters' behaviors, recovery costs, and service levels. Mathematical models are devised to compute monetary benefit functions for bus-only, taxi-only, and hybrid replacement services, with numerical analyses guiding the determination of the service type to implement. However, the study failed to establish a reliable long-term collaborative relationship between taxi, bus, and tram operators to cope with unpredictable disruptions. 

Studies on bridging services predominantly aim to reduce the bridging time and passenger delays, often under the assumption of an available dedicated reserve fleet, which may not align with the actual availability in real-world scenarios. In addition, the financial implications for operators have not been sufficiently explored. Existing studies primarily concentrate on decision-making models that are passenger-centric, often neglecting the considerations and needs of the operators, which does not accurately represent the complexities of systems with multiple concurrent operators. Moreover, the potential impact of deploying replacement vehicles on the surrounding road traffic during rail disruptions is frequently disregarded. There is also a notable deficiency in the research on the utilization of multimodal replacement fleets and the effective use of existing resources to address disruptions.
Recognizing these limitations, we introduced the RaaS concept, which is designed to address them. A detailed enumeration of the contributions of RaaS is presented in the subsequent subsection.

\subsection{Contribution statement}

The RaaS model is distinguished from existing literature through several innovative contributions:\\
(i)	Multimodal Resource Optimization: Unlike traditional approaches, which focus on single-mode solutions, RaaS presents an integrated multimodal optimization framework. Considering all the available modes (buses, taxis, and automated vans) for the replacement process, it offers a flexible approach to disruption management. Additionally, we integrated automated vans, further expanding the set of available vehicles for the replacement process used for the first time within a disruption management context. \\
(ii) Dynamic Reallocation of Resources: RaaS dynamically reallocates resources in disruptive events, ensuring a more responsive and efficient management of disruptions and adaptation to the evolving needs of the transportation network.\\
(iii) Collaborative framework between service providers: The RaaS model promotes collaboration among various mobility services providers, such as buses, taxis, and automated vans, reflecting a more realistic approach to managing multimodal networks where multiple operators work concurrently. \\
(iv)	Multi-stakeholder impact analysis in disruption management: RaaS adopts a multi-stakeholder perspective that considers the roles and impacts of all parties involved, including the main service provider, auxiliary service providers, and passengers. This inclusive approach highlights the interconnected nature of transportation systems and ensures a deeper understanding of disruption management and its repercussions across the entire transportation network. \\
(v)	 Multi-cost–benefit analysis strategy: The cost-benefit analysis strategy of RaaS tackles disruptions by considering both operational costs from the service providers' standpoint and passengers' time costs, accounting for both their willingness to leave the disrupted station and their willingness to wait for alternative services established by the operator, unlike conventional methods, which often focus solely on one aspect. This nuanced approach provides insights into the interplay between the operator and commuter decisions.

\section{Problem statement}

Unexpected disruptions in rail systems can partially or completely stop mobility services between stations, leaving passengers stranded. The service provider must decide between a do-nothing approach, which can harm the operator’s reputation and customer satisfaction, or alternative transport modes to assist blocked passengers and reduce losses \cite{zeng2012collaboration}. The do-nothing strategy refers to when the operator does not establish any replacement service for blocked passengers, leaving them to find alternative solutions independently. This approach is characterized by a complete loss of passenger loyalty, according to \cite{fang2020disruption}. Meanwhile, passengers waiting at disrupted stations must decide whether to leave the station or wait for the initiation of a replacement service. This scenario presents a complex challenge requiring a strategic balance between operational efficiency and customer satisfaction. To define the scope of the disruptions we aim to manage, we consider a scenario in which a critical failure occurs along a transit line, affecting several consecutive stations. Such disruptions impact not only a single origin-destination pair but also a series of connections integral to the transportation network. Our model addresses these multi-node disruptions by dynamically reallocating resources, ensuring that all affected nodes are considered. 

To address such cases, this study proposes an optimization-based framework that integrates alternative vehicles from a variety of transport providers to mitigate service interruptions, which are often caused by unexpected events such as train malfunctions, leading to a reduction in service level and impacting passengers for durations typically ranging from one to a few hours. The conventional response of redirecting passengers from a disrupted station to the nearest operational station \citep{ fang2020disruption} is enhanced in our approach by optimizing resource reallocation and minimizing the arrival duration of replacement vehicles as well as the associated costs, both monetary and temporal. Our approach also incorporates passenger behavior, particularly their willingness to wait for or refuse to use established replacement services, as discussed by \cite{zeng2012collaboration}, while if disruption duration $TD^n_{ij}$ is a known parameter. The overarching goal is to reduce loyalty and monetary losses from the operator’s perspective and time loss from the passenger’s perspective, ultimately improving passenger satisfaction and alleviating financial strain on operators. This comprehensive strategy ensures minimal impact on the replacement modes and maintains service continuity during disruptions. 

In this replacement strategy, we engaged three principal stakeholders: the primary operator of the disrupted line, auxiliary operators tasked with rerouting vehicles to accommodate passengers from the obstructed main station and the passengers. In this study, the RaaS provider is integrated with the primary service provider. From the operator’s perspective, we established two key objectives that must be met to achieve the optimization model. First, we aim to minimize the monetary cost, which is equivalent to the monetary payment made by the operator to the collaborative companies. Second, we aim to minimize loyalty loss experienced by the operator by providing a fast replacement service to reduce passengers' willingness to leave the disrupted station, thereby minimizing wait duration and increasing the number of passengers served. The cost implications of disruptions caused by trip cancellations at the stations from which the replacement vehicles were coming were also considered. The RaaS commitment seeks an optimal solution to effectively achieve this dual objective. However, the inherent contradiction between these two goals necessitates the use of advanced optimization techniques.

\section{Problem formulation and assumptions}

This section presents a framework to address the challenges outlined in the problem statement. It begins by introducing the notations used throughout the proposed framework and presents an optimization model that considers primary and auxiliary service providers, including RaaS, and passengers as key stakeholders with distinct objectives. Service providers aim to minimize monetary costs and preserve customer loyalty, whereas passengers seek alternative transport modes while reducing their wait duration. These dual objectives guided the development of our framework, which was designed to determine the most effective replacement solution. We begin with an overview of the entire framework, highlighting the interplay between the various constraints and parameters that influence the decision-making process. Subsequent sections provide an in-depth examination of each element, including the network and disruption model, resource reallocation parameters, volume of affected passengers, and average arrival duration of replacement vehicles. Additionally, we consider passenger behavior by analyzing the departure rate and detailing the cost aspects of replacement services, including transferring passengers and arranging vehicle costs, which are essential for evaluating economic implications.

\begin{longtable}{p{1.5cm}p{14cm}}\hline
\caption{\textbf{Table of notations}} \\ \hline
\multicolumn{2}{l}{\text{Input Parameters}} \\
$S$ & Set of all stations.\\
$Cl$ & Cost of leaving per passenger.\\
$CT$ & Cost of time.\\
$CO^m$ & Operation cost of mode $\textit{m} \in M$.\\
$P_{max}$ & Maximum payment rate.\\
$P_{min}$ & Minimum payment rate.\\
$c^m$ & Capacity of a vehicle in mode $\textit{m} \in M$.\\
$s^m$ & Average speed of a vehicle in mode $\textit{m} \in M$.\\
$\alpha$ & Lowest level of leaving passengers at disrupted station.\\
$\beta$ & lowest level of waiting passengers at disrupted station.\\
$TA^{m,x}_{ij}$ & Arrival duration of vehicle \textit{x} to disrupted station, $\textit{x} \in X^m, \textit{m} \in M, \textit{i} \in I^n, \textit{j} \in J^n$.\\ 
$CA^{m,x}$ & Rate of arranging one vehicle in replacement mode \textit{m}, $\textit{x} \in X^m, \textit{m} \in M$.\\
$H_{max}$ & Headway threshold.\\ \hline

\multicolumn{2}{l}{\text{Disruption Parameters}} \\
$N$ & Set of disrupted modes.\\
$I^n$ & Set of disrupted stations for disrupted mode, $\textit{n} \in N, I^n \subset S$.\\
$J^n$ & Set of destinations having origins $\textit{i} \in I^n, \textit{n} \in N, J^n \subset S$.\\
$\bar I^n$ & Set of non-disrupted stations for disrupted mode $\textit{n} \in N$, $\bar I^n \subset S$.\\
$\bar J^n$ & Set of destinations having origins $\textit{i} \in \bar I^n$, $\textit{n} \in N, \bar J^n \subset S$.\\
$OD^n$ & Matrix of all origin-destination pairs \textit{(i,j)} for disrupted mode \textit{n}, $\textit{n} \in N, \textit{i} \in I^n, \textit{j} \in J^n$.\\
$TD^n_{ij}$ & Disruption duration, $\textit{i} \in I^n, \textit{j} \in J^n, \textit{n} \in N$.\\
$V^n_{ij}$ & Aggregated total volume of blocked passengers for disrupted mode \textit{n} going from \textit{i} to \textit{j}, $\textit{n} \in N, \textit{i} \in I^n, \textit{j} \in J^n$.\\ 
$\textit{i}$ & Index representing an origin station within the set of disrupted stations $I^n$ for a disrupted mode $\textit{n} \in N$. \\ 
$\textit{j}$ & Index representing a destination station within the set of destinations $J^n$ for a disrupted mode $\textit{n} \in N$.\\ \hline
\multicolumn{2}{l}{\text{Replacement service Parameters}} \\
$M$ & Set of available replacement modes.\\
$X^m$ & set of vehicles in mode $m \in M$.\\
$I^m$ & Set of disrupted stations for non-disrupted mode $m \in M$, $I^m \subset S$.\\
$J^m$ & Set of destinations having origins $i \in I^m, \textit{m} \in M, J^m \subset S$.\\
$\bar I^m$ & Set of non-disrupted stations for non-disrupted mode $\textit{m} \in M$, $\bar I^m \subset S$.\\
$\bar J^m$ & Set of destinations having origins $\textit{i} \in \bar I^m$, $\textit{m} \in M, \bar J^m \subset S$.\\
$V^m_{rs}$ & Volume of blocked passengers at deliberately disrupted station \textit{r} going to \textit{s} by mode \textit{m}, $\textit{m} \in M, \textit{r} \in I^m, \textit{j} \in J^m$.\\
$H^m_{rs}$ & Headway of the line of mode \textit{m} serving link \textit{(r,s)}, $\textit{m} \in M, \textit{r} \in \bar I^m, \textit{s} \in \bar J^m$. \\
$\textit{r}$ & Index representing an origin station within the set of non-disrupted stations $\bar{I}^m$ for a non-disrupted mode $\textit{m} \in M$. \\ 
$\textit{s}$ & Index representing a destination station within the set of destinations $\bar{J}^m$ for a non-disrupted mode $\textit{m} \in M$.\\ \hline

\multicolumn{2}{l}{\text{Calculated Parameters}} \\
${TA^m_{ij}}$ & Aggregated average arrival duration to disrupted station, $\textit{m} \in M, \textit{i} \in I^n, \textit{j} \in J^n$.\\
$L_{ij}$ & Aggregate passenger’s departure rate, $\textit{i} \in I^n, \textit{j} \in J^n$.\\
$VL^n_{ij}$ & Volume of leaving passengers at the main disrupted station, $\textit{n} \in N, \textit{i} \in I^n, \textit{j} \in J^n$.\\
$VW^n_{ij}$ & Volume of waiting passengers at the main disrupted station, but not served by the replacement process $\textit{n} \in N, \textit{i} \in I^n, \textit{j} \in J^n$.\\
$VL^m_{rs}$ & Volume of leaving passengers at deliberately disrupted station, $\textit{m} \in M, \textit{r} \in I^m, \textit{s} \in J^m$.\\
$VW^m_{rs}$ & Volume of waiting passengers at deliberately disrupted station $\textit{m} \in M, \textit{r} \in I^m, \textit{s} \in J^m$.\\
$CS^m$ & Cost of service per mode $\textit{m} \in M$.\\
$CA(\gamma^{rs,m,x}_{ij})$ & Aggregate cost of arranging vehicles, $\textit{r} \in \bar I^m, \textit{s} \in \bar J^m, \textit{m} \in M, \textit{i} \in I^n, \textit{j} \in J^n$.\\
$U^m_{ij}$ & Number of used replacement vehicles in mode \textit{m}, $\textit{m} \in M, \textit{i} \in I^n, \textit{j} \in J^n$.\\
$d^{m,x}_{ri}$ & Travel distance of vehicle \textit{x} of mode \textit{m} from its initial location \textit{r} to the disrupted station \textit{i}, $\textit{m} \in M, \textit{x} \in X^m, \textit{r} \in \bar I^m, \textit{i} \in I^n$.\\
$d^{m,x}_{ij}$ & Travel distance of vehicle \textit{x} of mode \textit{m} from the disrupted station \textit{i} to the destination \textit{j}, $\textit{m} \in M, \textit{x} \in X^n, \textit{i} \in I^n, \textit{j} \in J^n$.\\ \hline

\multicolumn{2}{l}{\text{Decision Parameters}} \\ 
$\pi^n_{ij}$ & Binary parameter to identify main disrupted station, $\textit{n} \in N, \textit{i} \in I^n, \textit{j} \in J^n$. \\
$\xi^m_{rs}$ & Binary parameter to identify deliberately disrupted station, $\textit{m} \in M, \textit{r} \in \bar I^m, \textit{s} \in \bar J^m$.\\
$\gamma^{rs,m,x}_{ij}$ & Binary parameter to identify the reallocation of vehicle \textit{x} planned to serve the link \textit{(r,s)} in mode \textit{m} to serve the link \textit{(i,j)} of disrupted mode \textit{n}, $\textit{r} \in \bar I^m, \textit{s} \in \bar J^m, \textit{m} \in M, \textit{x} \in X^m, \textit{i} \in I^n, \textit{j} \in J^n$. \\ \hline
\end{longtable}

\subsection{Network modeling}
The transport network consists of a collection of $K$ encompassing all the modes of transportation within the system. During a disruption, the network is divided into two groups: a set of modes affected by the disruption, denoted by $N \subset K$, and a set of modes available as alternatives, denoted by $M \subset K$. Represent the multimodal network as a general directed graph $G^k = (S^k, L^k)$ for each mode $k \in K$, where $S^k$ is the set of all stations or stops and $L^k$ is the set of links connecting these stations. Station set $S^k$ is further categorized into four subsets: disrupted stations $I^k$, destinations with disrupted origins in $I^k$ denoted by $J^k$, nondisrupted stations $\bar{I}^k$, and destinations with nondisrupted origins denoted by $\bar{J}^k$. The composition of station set $S^k$ is represented by Equation (\ref{Sk}):

\begin{equation}
    S^k \label{Sk} = I^k \cup \bar{I}^k \cup J^k \cup \bar{J}^k 
\end{equation}

To distinguish between stations affected by disruptions and those that were not, we assigned the index $n$ to disrupted stations and $m$ to unaffected stations. During a disruption, the affected mode $n \in N$ operating at $G^n=(S^n,L^n)$ will have disrupted stations/stops and a certain number of links connecting these stations to their destinations. The set of disrupted stations is represented by $I^n$, whereas the set of nondisrupted stations is represented by $\bar{I}^n$. The set of all destinations with at least one disrupted origin $I^n$ is denoted by $J^n$, and the set of all destinations with a nondisrupted origin $\bar{I}^n$ is denoted by $\bar{J}^n$. 
Similarly, the alternative modes $m \in M$ operating on $G^m=(S^m,L^m)$ have disrupted stations, represented by $I^m$, along with their destination set $J^m$, and nondisrupted stations, represented by $\bar{I}^m$, along with their destination set $\bar{J}^m$.
During the replacement process, we assume $OD^n$ is the matrix of origin-destination pairs for the disrupted mode $n \in N$, with indices $i \in I^n$ and $j \in J^n$, and $X^m$ is the set of all available vehicles that can be utilized in the replacement process.

\subsection{Decision parameters}

\subsubsection{Disruption parameter}

We consider that disruptions occur within the public transportation modes denoted by $\textit{n} \in N$, whereas all other operational modes indicated by $\textit{m} \in M$ may assist passengers at disrupted stations. $\pi^n_{ij}$ is an input parameter to the model that denotes the status of mode \textit{n} on link \textit{(i,j)}, with a value of zero indicating no disruption and a value of one signifying a disruption. This parameter helps to identify disrupted stations. If $\pi^n_{ij}$ equals one, then station \textit{i} is classified as disrupted and included in set $I^n$, and mode \textit{n} is added to the set of disrupted modes $N$. This parameter is defined as follows:

\begin{equation} \label{pi}
     \pi^n_{ij} =  \begin{cases}
    1 & \text{If mode \textit{n} serving link \textit{(i,j)} is disrupted} \\
    0 & \text{Otherwise}
    \end{cases} \quad
\end{equation}

\subsubsection{Resources reallocation parameters}

The objective of RaaS is to reallocate specific vehicles to disrupted stations. This reallocation occurs either at the nearest stations or depots for vehicles belonging to public transport modes, such as buses, or at the closest locations for modes such as taxis, depending on cost-effectiveness. The process involves redirecting vehicles to alternate routes to fulfill the needs of the affected passengers. Additionally, we consider the concept of deliberately disrupted stations. When a vehicle is selected to participate in the replacement process and has previously been serving link \textit{(r,s)} on another line, the station \textit{r} associated with that link must be declared disrupted. 

The binary variable $\gamma^{rs,m,x}_{ij}$ denotes the reallocation of vehicle \textit{x} of type \textit{m}, originally intended to serve link $\textit{(r,s)}$ to assist link $\textit{(i,j)}$ of the disrupted mode $n$. If vehicle \textit{x} is serving link $\textit{(r,s)}$ without supporting link $\textit{(i,j)}$, $\gamma^{rs,m,x}_{ij}$ is zero. Conversely, if vehicle \textit{x} is redirected to support the disrupted service on link $\textit{(i,j)}$, $\gamma^{rs,m,x}_{ij}$ is one. 
The decision parameters are defined as follows.

\begin{equation}
    \gamma^{rs,m,x}_{ij} =  \begin{cases}
    1 & \text{if vehicle \textit{x} in mode \textit{m} planned to serve the link \textit{(r,s)} is reallocated to support the link \textit{(i,j)} having disrupted service}  \\
    0 & \text{otherwise}
    \end{cases} \quad
\end{equation}

This decision parameter determines the vehicles to be rerouted to mitigate disruptions. The total number of vehicles used in the process equals the sum of all the assigned vehicles, as shown in Equation (\ref{c3}). The flexibility in defining this variable enables a versatile model to handle various transport modes and scenarios. 

To identify deliberately disrupted stations, we use the binary variable $\xi^m_{rs}$ to indicate whether mode $m$ is deliberately disrupted for a given link, \textit{(r,s)}. The value of $\xi^m_{rs}$ is zero when mode \textit{m} normally serves its initial link \textit{(r,s)}, and one when we deliberately disrupt it to serve another link. These parameters are defined as follows:

\begin{equation} \label{xi} 
\xi^m_{rs} = \begin{cases} 1 & \text{If the mode \textit{m} planned to serve the link \textit{(r,s)} is deliberately disrupted to serve a disrupted link} ;\\ 0 & \text{Otherwise} \end{cases} \quad 
\end{equation}

\subsection{Optimization model}

The optimization model was designed to implement the RaaS strategy. Its objective function is twofold: first, to minimize the operator’s monetary cost (Z1), including the primary and auxiliary operators, as detailed in Equation (\ref{Z1}); second, to minimize the loyalty cost of dissatisfied passengers, as outlined in Equation (\ref{Z2}). In this section, we discuss the specifics of the objective function presented in Equation (\ref{OF}), along with the constraints and parameters that shape the optimization model. The model was formulated as a MILP problem that was configured to be solvable using standard commercial optimization software. However, it is also designed flexibly, allowing adaptation to a variety of scenarios without the need for heuristic methods. This ensures that the model remains applicable across different operational contexts, providing a reliable tool for decision-making during service disruptions.
\begin{align} \label{OF}
\min_{\gamma^{rs,m,x}_{ij}} \enspace Z_1(\gamma) + Z_2(\pi,\xi) \\
s.t. \quad 
& \pi^n_{ij} + \xi^m_{rs} \le 1 \label{c1} \quad \forall n \in N, \forall i \in I^n, \forall j \in J^n, \forall r \in \bar{I}^m, \forall s \in \bar{J}^m, \forall m \in M\\
& (1- \xi^m_{rs})\sum_{i \in I^n} \sum_{j \in J^n} \sum_{x \in X^m} \gamma^{rs,m,x}_{ij}  = 0 \label{c2} \quad \forall r \in \bar{I}^m, \forall s \in \bar{J}^m, \forall m \in M\\ 
& U^m_{ij} = \sum_{r \in \Bar{I}^m} \sum_{s \in \Bar{J}^m} \sum_{x \in X^m} \gamma^{rs,m,x}_{ij} \label{c3} \quad \forall i \in I^n, \forall j \in J^n, \forall m \in M\\
& \xi^m_{rs} \cdot H^m_{rs} \leq H_{max} \label{c4} \quad \forall r \in \bar{I}^m, \forall s \in \bar{J}^m (s \ne r), \forall m \in M\\
&  {TA^m_{ij}} <= TD^n_{ij} \label{c5} \quad \forall i \in I^n, \forall j \in J^n, \forall m \in M, \forall n \in N
\end{align}

Where $Z_1$ and $Z_2$ are calculated as 

\begin{equation} 
Z_1(\gamma) \label{Z1} = \sum_{i \in I^n} \sum_{j \in J^n} \sum_{r \in \bar{I}^m} \sum_{s \in \bar{J}^m} \sum_{m \in M} \sum_{x \in X^m} [\gamma^{rs,m,x}_{ij} \cdot (CS^m \cdot c^m \cdot d^{m,x}) + CA(\gamma^{rs,m,x}_{ij})]
\end{equation}

%\begin{equation}
%Z_2(\pi,\xi) \label{Z2} = \sum_{i \in I^n} \sum_{j \in J^n} \sum_{n \in N} \pi^n_{ij} \cdot [(CL + (TD^n_{ij} %\cdot CT)) \cdot VL^n_{ij} + (TD^n_{ij} \cdot CT) \cdot VW^n_{ij})] + \sum_{r \in \bar{I}^m} \sum_{s \in %\bar{J}^m} \sum_{m \in M} \xi^m_{rs} \cdot [(CL + (H^m_{rs} \cdot CT)) \cdot VL^m_{rs} + (H^m_{rs} \cdot CT) %\cdot VW^m_{rs}]
%\end{equation}

\begin{align}
Z_2(\pi,\xi) \label{Z2} = & \sum_{i \in I^n} \sum_{j \in J^n} \sum_{n \in N} \pi^n_{ij} \cdot [(CL + (TD^n_{ij} \cdot CT)) \cdot VL^n_{ij}
+ (TD^n_{ij} \cdot CT) \cdot VW^n_{ij})] + \\ & \sum_{r \in \bar{I}^m} \sum_{s \in \bar{J}^m} \sum_{m \in M} \xi^m_{rs} \cdot [(CL + (H^m_{rs} \cdot CT)) \cdot VL^m_{rs} + (H^m_{rs} \cdot CT) \cdot VW^m_{rs}] \nonumber
\end{align}

\subsection{Model constraints optimization}

This model integrates various constraints to enhance the efficient management of disruptions. First, constraint (\ref{c1}) prohibits sourcing replacement vehicles from already disrupted lines, thereby preventing deliberate disruptions at stations that have already been impacted. Second, constraint (\ref{c2}) ensures that any mode $m$ reallocated to serve link $\textit{(i,j)}$ of disrupted mode $n$ instead of its initial link $\textit{(r,s)}$, is disrupted by setting $\xi^m_{rs} = 1$. Constraint (\ref{c4}) imposes limitations on selecting in-service lines based on their headway duration, with a maximum headway $H_{max}$ set at the start of the disruption to regulate the passenger wait duration at deliberately affected stations. This constraint was specifically applied to public transport replacement vehicles to ensure that lines with relatively long headways were not tasked with assistance. Constraint (\ref{c3}) tracks the total number of vehicles utilized for a specific mode \textit{m}, considering the optimized count of reallocated vehicles once the replacement process concludes. Finally, Constraint (\ref{c5}) guarantees that the replacement vehicles arrive within the maximum tolerated wait duration.

\subsection{Replacement vehicles’ average arrival duration} 

The arrival duration of the replacement vehicles is a critical factor in the replacement process and is influenced by various elements, including the start time of the disruption and the availability of alternative transportation modes. 
${TA^m_{ij}}$ represents the average arrival duration of vehicles in substitute mode $m \in M$ at the disruption station to replace the mobility service for $\textit{(i,j)}$. We incorporated the average arrival duration for each replacement mode ${TA^m_{ij}}$ as an input parameter calculated using Equation (\ref{A}). The replacement vehicles originate from diverse locations and have varying arrival durations. ${TA^m_{ij}}$ is calculated as the sum of the arrival durations of each vehicle in mode $m$ divided by the number of vehicles used from mode $U^m_{ij}$ defined in Equation (\ref{c3}).
To ensure the effective operation of the replacement process, the average arrival duration of replacement vehicles mustn't exceed the disruption duration, as mentioned in Equation (\ref{T}).

\begin{equation}
     {TA^m_{ij}} \label{A} = \frac{\sum_{r \in \bar{I}^m} \sum_{s \in \Bar{J}^m}\sum_{x \in X^m}{TA^{m,x}_{ij}}.\gamma^{rs,m,x}_{ij}}{U^m_{ij}} \quad \forall m \in M, \forall i \in I^n, \forall j \in J^n, \forall r \in \bar{I^m}, \forall s \in \bar{J^m}, \forall x \in X^m\\ 
\end{equation}

\begin{equation} \label{T}
    0 \le {TA^m_{ij}} \le TD^n_{ij} \quad \forall m \in M, \forall n \in I^n, \forall i \in I^n, \forall j \in J^n \\ 
\end{equation}

Where $TA^{m,x}_{ij}$ is the arrival duration of a vehicle \textit{x} of type \textit{m}, $(x \in X^m)$, calculated as the traveled distance of the vehicle from its origin location $r$ to the disrupted station $i$, $d^{m,x}_{ri}$, divided by speed $s^m$, as shown in Equation \ref{Amx}, assuming that this arrival duration includes the congestion delay. 

\begin{equation}
    TA^{m,x}_{ij} \label{Amx} = \frac{d^{m,x}_{ri}}{s^m} \quad \forall m \in M, \forall x \in X^m, \forall r \in \bar{I}^m, \forall i \in I^n, \forall j \in J^n \\
\end{equation}

\subsection{Monetary cost calculation}

In the proposed model, we account for the service cost of the disrupted operator to replace companies transporting stranded passengers. To simplify the calculation process and focus on the direct costs from the operator's perspective, we assume that the passengers will not receive any refund, as well as they are not obliged to pay for other transport modes, as this cost oversees the service provider. The monetary cost, as outlined in Equation (\ref{Z1}), comprises two main components when vehicle \textit{x} of mode \textit{m} is reallocated to serve the line \textit{(i,j)} instead of its initial origin-destination trajectory \textit{(r,s)}: the payment for completed service, based on the service cost of mode \textit{m}, the distance traveled by each vehicle and its capacity, following a variable payment rate strategy, $(CS^m \cdot c^m \cdot d^{m,x})$, and the arrangement cost incurred for each demand $CA(\gamma^{rs,m,x}_{ij})$ calculated as in Equation (\ref{CA}). 

Calculating the distance traveled by each replacement vehicle, as in Equation (\ref{d}), requires consideration of the distance between its starting point \textit{r} and the final destination \textit{j}. This includes not only the distance covered between the origin and destination \textit{(i,j)} trajectory of the disrupted mode \textit{n} but also the distance covered before reaching the disrupted station. For instance, if the replacement vehicle belongs to a public transport system, such as a bus, we calculate the travel distance by measuring the distance between its initial station \textit{r} and disrupted station \textit{i}, and then from \textit{i} to destination \textit{j}. If the replacement vehicle originates from the depot, both \textit{r} and \textit{s} refer to the same location at the depot, whereas for vehicular replacement modes, such as taxis, \textit{r} refers to the original location from which it originates.

\begin{equation} \label{d}
    d^{m,x} = d^{m,x}_{ri} + d^{m,x}_{ij}  \quad \forall m \in M, \forall x \in X^m, \forall r \in \bar{I}^m, \forall i \in I^n, \forall j \in J^n
\end{equation}

We excluded the distance traveled beyond destination \textit{j} from this calculation because the subsequent destination of the rerouted vehicle \textit{x} is beyond the considered disruption duration; it may return to its initial origin \textit{r} or destination \textit{s} (e.g., as a backup fleet or bus) or serve elsewhere in the network (e.g., taxi).

\subsection{Replacement service cost}

Replacement costs involve expenses associated with securing an alternative mode of transport to substitute for a disrupted transport mode along a specific link. The magnitude of this cost varies, depending on the severity of the disruption and the mode of transport affected. In our model, the replacement cost comprises two primary components: the cost of transferring passengers to destination stations, which is influenced by the variable cost of the service strategy, and the cost of arranging suitable vehicles.

\subsubsection{Transferring cost}

To motivate the replacement companies to provide fast services, the payment rate was adjusted based on the arrival duration ${TA^{m,x}_{ij}}$ of each replacement vehicle. Specifically, the payment amount decreases as the arrival duration increases, reflecting the notion that slower replacement processes lead to lower payments \citep{fang2019replacement}. Let $CS^{m,x}$ represent the variable cost of service (in €/passenger.km) for vehicle \textit{x} of type \textit{m} provided by companies involved in transferring stranded passengers during a disruption. Let $P_{max}$ be the maximum payment rate for replacement vehicles arriving within the first half of the disruption before $\frac{TD^n_{ij}}{2}$ and $P_{min}$ be the minimum payment rate for replacement vehicles arriving in the second half of the disruption. If replacement vehicles arrive after the disruption ends, y will assume the value $-\frac{P_{min}}{P_{max} - P_{min}}$ based on Equation (\ref{y}), and $CS^{m,x}$ will have the value of zero, implying that only an arrangement cost $CA(\gamma^{rs,m,x}_{ij})$ will be paid for arranged vehicles in this case. Based on this payment strategy, the cost of service for a given replacement mode $m$, $CS^{m,x}$, is calculated using Equation \ref{CS}.

\begin{equation} \label{CS}
CS^{m,x} = CO^m \cdot [y \cdot P_{max} + (1 -y)P_{min}] \quad \forall m \in M, \forall x \in X^m \\ 
\end{equation}

Where $CO^m$ represents the operational cost for replacement mode $m$ (in €/passenger.km), and $y$ is a piecewise function indicating the payment variation based on the arrival duration ${TA^{m,x}_{ij}}$ of each replacement vehicle.

\begin{equation} \label{y}
y = \begin{cases} 1 \;\;\;\;\;\;\;\;\; \; \;\;      {{TA^{m,x}_{ij}}} \le \frac{TD^n_{ij}}{2} \\ 0 \;\;\;\;\;\;\;\;   \frac{TD^n_{ij}}{2} < {{TA^{m,x}_{ij}}} < TD^n_{ij}\\ -\frac{P_{min}}{P_{max} - P_{min}} \;\;\;\;\;\;\;\;  {{TA^{m,x}_{ij}}} \ge TD^n_{ij} \end{cases} \quad \\
\end{equation}

\subsubsection{Arrangement cost}

Organizing the available replacement vehicles presents a significant challenge for companies involved in the replacement process. This task includes internal communication, determination of vehicle location, driver assignment, ensuring sufficient capacity, and scheduling. The cost of arranging replacement vehicles is a function of both the vehicle reallocation parameter $\gamma^{rs,m,x}_{ij}$ and arrival duration of vehicles ${TA^{m,x}_{ij}}$. The associated cost reflects the fact that the effort and expenses escalate more rapidly as the requested arrival duration shortens, as discussed in \cite{zhang2020metro}. This cost is calculated using \ref{CA}.

\begin{equation} \label{CA}
    CA(\gamma^{rs,m,x}_{ij}) = \frac{CA^{m,x} \cdot \gamma^{rs,m,x}_{ij}}{TA^{m,x}_{ij}} \quad \forall i \in I^n, \forall j \in J^n, \forall m \in M, \forall x \in X^m \\
\end{equation}
Where $CA^{m,x}$ is the rate of arranging one vehicle \textit{x} in mode \textit{m} ( €/unit time). Equation (\ref{CA}) models the arrangement cost as inversely proportional to the arrival duration of the replacement vehicles $TA^{m,x}_{ij}$. As $TA^{m,x}_{ij}$ decreases, indicating a need for more rapid deployment of replacement vehicles, the value of $\frac{1}{TA^{m,x}_{ij}}$ increases, thereby increasing the overall arrangement cost $CA(\gamma^{rs,m,x}_{ij})$ as the values of $CA^{m,x}$ and $\gamma^{rs,m,x}_{ij}$ are positive.

\subsection{Loyalty cost calculation}

To compute the cost of loyalty loss from the operator's perspective, we categorized passengers into two groups: those who leave the disrupted station without waiting for the substitute process and those who wait at the disrupted station without being served by the replacement vehicles owing to lack of capacity. These loyalty loss costs are calculated for disruptions primarily at disrupted stations and deliberately disrupted stations in Equation (\ref{Z2}), involving two primary components:
(i)	 Cost incurred by departing passengers, calculated as $(CL + (TD^n_{ij} \cdot CT)) \cdot VL^n_{ij})$ for primarily disrupted stations and $(CL + (H^m_{rs} \cdot CT)) \cdot VL^m_{rs})$ for deliberately disrupted stations. Here, $CL$ represents the cost of leaving a passenger, $CT$ signifies the time cost per passenger, and $VL^n_{ij}$ and $VL^m_{rs}$ denote the volumes of departing passengers at the primary and deliberately disrupted stations, respectively.
(ii) The cost of passengers waiting without service, expressed as $(TD^n_{ij} \cdot CT) \cdot VW^n_{ij})$ for primarily disrupted stations and $(H^m_{rs} \cdot CT) \cdot VW^m_{rs})$ for deliberately disrupted stations, where $VW^n_{ij}$ and $VW^m_{rs}$ represent the volume of waiting passengers. 

The cost of time \textit{CT} is included because when passengers decide to leave a disrupted station or wait for a replacement service to arrive, they incur a loss in terms of time. This is a direct cost to them and, by extension, a cost to the operator in terms of lost loyalty, that is, from the operator’s perspective, long-term revenue loss due to passengers potentially abandoning the service. By considering the time cost multiplied by the disruption duration $TD^n_{ij}$ for disrupted stations, or headways $H^m_{rs}$ for deliberately disrupted stations, the operator can quantify the potential long-term impact of disruptive events on its reputation. 

The number of passengers leaving the disrupted station denoted by $VL^n_{ij}$, was determined by multiplying the departure rate $L_{ij}$ calculated in Equation (\ref{L}) by the total number of stranded passengers $V^n_{ij}$, as expressed in Equation (\ref{VLn}). To ascertain the volume of waiting for passengers not served by the replacement service, denoted by $VW^n_{ij}$, it is necessary to subtract the volume of passengers leaving $VL^n_{ij}$ from the number of passengers served by the replacement service $\sum_{m \in M}{U^m_{ij}.c^m}$ from the total volume of blocked passengers $V^n_{ij}$, as shown in Equation \ref{VWn}. 
\begin{equation} \label{VLn}
VL^n_{ij} = L_{ij} \cdot V^n_{ij} \quad \forall i \in I^n, \forall j \in J^n \\
\end{equation}

\begin{equation} \label{VWn}
VW^n_{ij} = max(0, V^n_{ij} - VL^n_{ij} - \sum_{m \in M}\sum_{x \in X^m}{\gamma^{m,x}_{ij}.c^m}) \quad \forall i \in I^n, \forall j \in J^n, \forall m \in M, \forall x \in X^m \\
\end{equation}

The calculations of $VL^n_{ij}$ and $VW^n_{ij}$ for deliberately disrupted stations follow a similar methodology, as shown in Equations \ref{VLm} and \ref{VWm}, respectively, under the assumption that no bridging process is implemented for these disruptions.

\begin{equation} \label{VLm}
VL^m_{rs} = L_{rs} \cdot V^m_{rs} \quad \forall r \in \bar{I}^m, \forall s \in \bar{J}^m 
\end{equation}

\begin{equation} \label{VWm}
VW^m_{rs} = max(0, V^m_{rs} - VL^m_{rs}) \quad \forall r \in \bar{I}^m, \forall s \in \bar{J}^m, \forall m \in M \\
\end{equation}

\subsection{Volume of affected passengers}

During unplanned disruptions, an accurate assessment of the number of stranded passengers is essential for calculating the number of vehicles that should be reallocated. This volume of affected passengers not only depends on the severity of the disruption and passengers’ reliance on the affected mode of transport but also varies based on the time of day when the disruption occurs.
For each origin-destination pair \textit{(i,j)}, the total number of blocked passengers throughout the disruption duration can be estimated using Equation (\ref{V}).
\begin{equation} \label{V}
V^n_{ij} = \int_{t \in TD^n_{ij}} D^n_{ij}(t)\,dt \quad \forall i \in I^n, \forall j \in J^n, \forall n \in N \\
\end{equation}

Where $D^n_{ij}(t)$ denotes the passenger demand for the origin-destination pair (i,j) at a given time t, which varies between the start and end times of the disruption. 

In our approach, we leverage the MATSim synthetic population data specifically created for the Île-de-France region by \cite{horl2021synthetic}. This synthetic population provides a representative sample of travelers, capturing their travel patterns, preferences, and behaviors based on the National Institute of Statistics and Economic Studies INSEE, and estimates the demand volume for the disrupted mode \textit{(n)} serving the link \textit{(i,j)} derived from the synthetic population. 

\subsection{Passengers’ departure rate}

Passenger response to disruptions, whether they opt to await replacement vehicles or leave the disrupted station, is contingent upon individual circumstances. However, the primary factor influencing their decisions was the wait duration for replacement vehicles. For those opting to wait, the wait duration is directly related to the arrival duration of the replacement vehicles at the disruption location. Based on the methodology proposed by \cite{zeng2012collaboration}, passenger willingness to wait was quantified using the following equation:

\begin{equation}
    W \label{W} = 1 - (1 - \theta) \cdot \frac{TA^m_{ij}}{TD^n_{ij}}
\end{equation}

Here, $\theta$ denotes the minimum level of passenger willingness to wait. The aggregate passenger willingness to wait is defined as the lowest tolerance level for waiting if replacement vehicles are expected to arrive significantly late. Inspired by Equation (\ref{W}), we develop a parameter to determine the departure rate of passengers from a disrupted station. Our model is based on the concept that there is a minimum number of passengers who will invariably leave the station immediately upon disruption \textit{$\alpha$} and a separate minimum number of passengers who will choose to wait for the replacement process, regardless of its duration \textit{$\beta$}.
Building on this concept, we define the aggregate departure rate $L_{ij}$ as the percentage of passengers who opt to leave the disrupted station \textit{i} instead of waiting for replacement vehicles. $L_{ij}$ was determined based on the average arrival duration of the replacement modes. Consequently, it increases as the arrival duration of the replacement vehicles increases, signifying an increase in the number of departing passengers. $L_{ij}$ was calculated using Equation (\ref{L}).

\begin{equation} \label{L}
    L_{ij} = \alpha + (1 - \beta - \alpha) \cdot \frac{{min_{m \in M}TA^{m,x}_{ij}}}{TD^n_{ij}} \quad \forall i \in I^n, \forall j \in J^n, \forall m \in M\\
\end{equation}

\newref{Figure}{Fig1} illustrates this equation, depicting the departure rate of passengers in response to the disruption. As $TA^m_{ij}$ is initiated at time zero, the departure rate remains at its minimum value $\alpha$. Over time, the departure rate gradually increases, reaching its peak value of $(1-\beta)$ when $TA^m_{ij}$ approaches $TD^n_{ij}$, indicating a significant delay in the replacement process and the highest value for leaving passengers.

\begin{figure}[h!]
\centering
\includegraphics[width=0.75\textwidth]{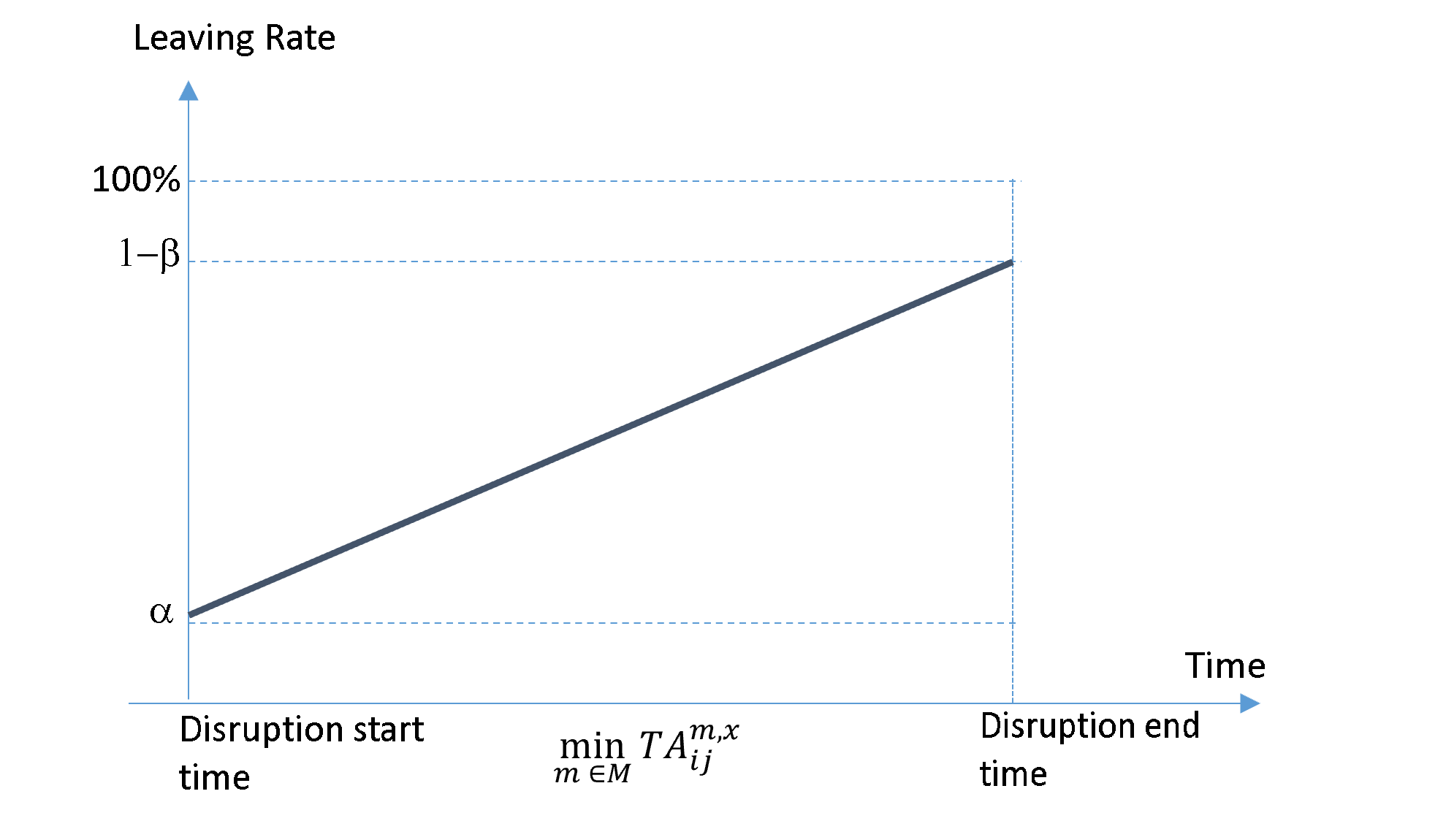}
\caption{The leaving rate of passengers in reaction to the disruption.}
\label{Fig1}
\end{figure}

\section{Methodology}

The methodological framework is visually depicted in \newref{Figure}{Figure2}, which illustrates the sequential flow of the methodology steps. It unfolds in six main steps.

\begin{figure}[h!]
\centering
\includegraphics[width=1\linewidth]{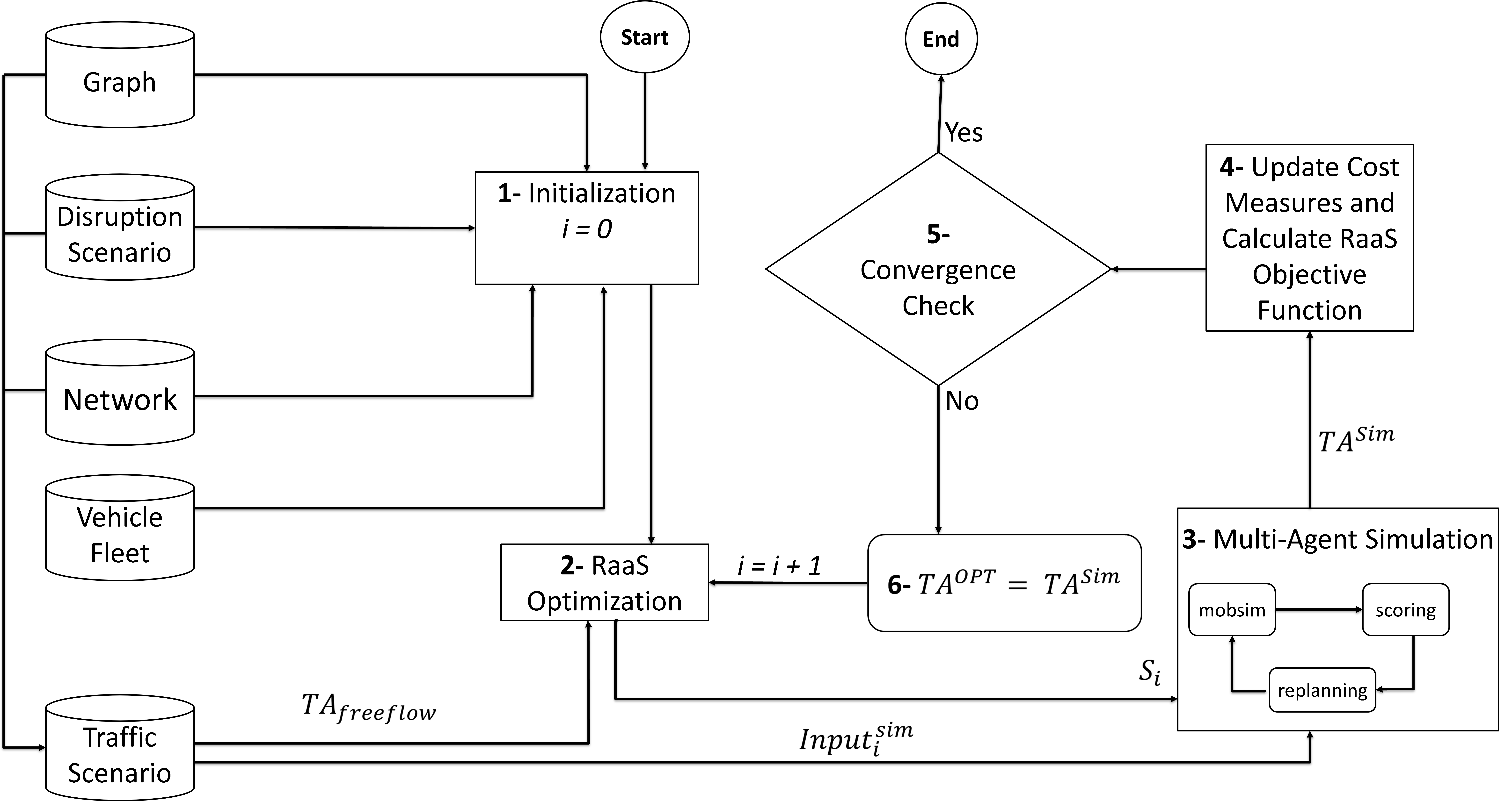}
\caption{RaaS dynamic methodological framework.}
\label{Figure2}
\end{figure}

\noindent \textbf{Step 1:} This step represents the foundation establishment for the entire model in the first iteration ($i = 0$). It begins with the collection and integration of data, including graphs, networks, scenarios, vehicle fleets, and demands, to create a traffic scenario. The data were then processed to create scenarios that served as baselines for the simulation. The initial conditions were set to define the state of the network and the starting points for all agents within the simulation. Key parameters such as passenger volume, disruption duration, and convergence criteria were determined, and the model was verified and validated. 

\noindent \textbf{Step 2:} The optimization loop is initiated with the initial average arrival duration of the replacement vehicles, which serves as a benchmark for subsequent improvements. The optimizer then engages in a systematic process of adjustments in which it evaluates the effectiveness of the current replacement vehicle deployment against the desired outcomes. If the cost and average arrival duration do not align with the optimal conditions, the optimizer recalculates the parameters, seeking to minimize them, and thus enhance the responsiveness of the transportation system to disruptions. This process continues with the optimizer making incremental changes and reassessing the results until the average arrival duration converges to a value within acceptable limits. The optimization process uses the initial free-flow travel duration ($TA_{\text{freeflow}}$) and generates input data for the simulation ($Input^{Sim}_i$). The shortest paths for the replacement vehicles were determined using Dijkstra’s algorithm, leveraging the comprehensive network data provided by MATSim. This approach ensures that the most efficient and effective routes are selected for vehicle deployment during disruptions. 

\noindent \textbf{Step 3:}  Using simulations, we explore the consequences of disruptions and test the effectiveness of different strategies for managing them. In this step, the demand, network, and vehicle files are inputted into a simulator that models the behavior of individual travelers and their interactions within the transportation network, providing key performance indicators (KPIs) and updating the objective function based on the simulation results. 
To achieve this objective, we employed MATSim, a mid-term agent-based simulation framework designed to handle large-scale scenarios that can simulate resource utilization in transportation. Its ability to simulate single- and multi-day scenarios adds to its versatility, making it a valuable tool for transportation planning and policy analysis \citep{ nguyen2021overview}. The framework operates on activity-based approach models of travel demand by assigning activities to agents, with the demand changing at each iteration owing to the replanning mechanism. The agents have multiple plans that encompass various actions and choices, including routes, travel modes, and time scheduling. Decisions are made using a discrete-choice model, in which agents select plans based on their scores; higher scores increase the likelihood of selection \citep{w2016multi}. 

\noindent \textbf{Step 4:} After the simulation, the cost measures are updated and the RaaS objective function is calculated based on these measures. 

\noindent \textbf{Step 5:} After each simulation iteration, a convergence check was performed to determine whether the optimized average arrival duration aligned with the simulated outcomes. If convergence is reached, which indicates that the system has stabilized and no further improvements can be made, the process ends. Otherwise, the speed of vehicle $S_i$ is updated, and the loop continues with further optimization. 

\noindent \textbf{Step 6:} Upon the completion of each iteration, the average arrival duration of the replacement vehicles ($TA^{Sim}$) is obtained from the simulation outcomes. This empirical value is then used to update the optimized average arrival duration ($TA^{OPT}$), which was previously determined using the optimization algorithm. The update process aligns the optimization model with the actual performance observed in the simulation, ensuring that the RaaS strategy remains grounded in practical conditions. Here, the iteration counter is incremented ($i = i + 1$), and the process may repeat from Step 2 if necessary.

\section{Case study and assumptions}

To evaluate the performance of the proposed model, we applied it to one of the most heavily utilized commuter train lines in the Paris metropolitan area, the RER B (Réseau Express Régional). The RER B line, which spans approximately 80 kilometers, connects the northern and southern suburbs through the heart of Paris, with 47 serving stations and approximately 983000 passengers per day. Our case study focuses on a fleet failure event occurring between two stations served by this line at 12 km, during the critical morning peak hours from 7:00 to 9:00 AM. 
The simulated region spans an approximate radius of 10 km from the affected station. During the morning peak hours from 7:00 to 9:00 AM, the demand volume within this area amounted to approximately 32,876 passengers, utilizing various public transportation modes, notably RER B and bus lines. This demographic information is sourced from the Île-de-France population synthetic dataset for MATSim, ensuring that every simulated passenger has a planned itinerary for a full 24-hour period. Our analysis focuses on the disruptions occurring during trips accessing RER B between 7:00 and 9:00 AM, which typically constitute home-to-work or home-to-education commutes, and their impact on the entirety of passengers' itineraries. Computations are conducted regarding travel duration, waiting duration, and distance covered, where the travel duration of a passenger for one trip $TT^{Passenger}_{Trip}$ encompasses both in-vehicle durations $IT^{Passenger}_{Trip}$ and walking durations $WT^{Passenger}_{Trip}$, as follows.

\begin{equation} \label{TT}
TT^{Passenger}_{Trip} = IT^{Passenger}_{Trip}+ WT^{Passenger}_{Trip}              
\end{equation}

By evaluating output parameters such as travel duration, waiting duration, and travel distance, alongside assessing resource allocation costs, we explored the system's performance across the following scenarios:
\begin{enumerate}
    \item Normal: Represents the absence of disruptive events affecting RER B services.
    \item Do-Nothing: The RER B service experiences disruption, compelling commuters to independently seek alternative transportation modes.
    \item RaaS: Involves implementing a substitute service under the RaaS strategy.
    \item Bus bridging: Bridging buses dispatched from the depot were introduced to mitigate the impact of disruptions.
    \item Taxi Bridging: Deploys bridging taxis to provide alternative transportation options.
    \item Automated Van bridging: Automated vans with a capacity of eight passengers each to offer supplementary transportation services. Although automated shuttles remain in experimental trials in Europe for analysis, we assume that they are commercially available to mobility operators.
\end{enumerate}

Scenarios 4, 5, and 6 are employed to benchmark our strategy in scenario 3 against the strategies used in \cite{ pender2013disruption}. 
We set the scenario parameters as detailed in \newref{Table}{table2}.

\begin{table}[!h]
\centering
\caption{Scenario parameters and assumptions}
\resizebox{\textwidth}{!}{\begin{tabular}{|l|p{0.6\linewidth}|}
\hline
\textbf{Parameter} & \textbf{Value} \\
\hline
Volume of blocked passengers at disrupted station ($V^n_{ij}$) & 300 \\
\hline
Headway threshold for replacement lines ($H_{max}$) & 15 minutes \\
\hline
Vehicle capacity per mode ($c^m$) & 
Bus: 70 \\
& Rail: 400 \\
& Subway: 300 \\
& Tram: 180 \\
& Taxi: 4 \\
& Automated van: 8 \\
\hline
Average speed per mode ($s^m$) & Variable based on MATSim network, depending on the free flow speed of each route. \\
\hline
Disruption duration ($TD^n_{ij}$) & 120 minutes \\
\hline
Minimum payment rate ($P_{min}$) & 0.3 (30\%) \\
\hline
Maximum payment rate ($P_{max}$) & 1 (100\%) \\
\hline
Arrangement rate per vehicle ($CA^{m,x}$) & 0.2 (20\% of the transfer cost based on travel distance). \\
\hline
Lowest level of leaving rate ($\alpha$) & 0.1 (10\%) \\
\hline
Lowest level of waiting rate ($\beta$) & 0.1 (10\%) \\
\hline
Cost of leaving (\textit{CL}) & 2.50 euros/passenger (Calculated as the full ticket price the disrupted station to the destination \citep{SNCF2024}). \\
\hline
Cost of time (\textit{CT}) & 11.2 euros/passenger.hour based on \citep{themamob2020}. \\
\hline
Operational cost per mode ($CO^m$) & 
Bus: 0.454 euros/passenger.km \\
& Rail: 0.139 euros/passenger.km \\
& Subway: 0.194 euros/passenger.km \\
& Tram: 0.196 euros/passenger.km \\
& Taxi: ((1.72 * TotalDistance) + 2.2) euros/passenger.km \\
& Automated van: 0.36 euros/passenger.km \citep{themamob2020, bosch2018cost, carreyre2023robotaxis}. \\
\hline
\end{tabular}}
\label{table2}
\end{table}
\clearpage

\section{Results and Discussion}

In this section, we present the results and organize them into three main sections. First, we validated our framework by comparing the RaaS strategy for both normal and do-nothing scenarios. Second, we benchmark our proposed solution against alternative strategies, namely bus bridging, taxi bridging, and automated van bridging. Finally, we conducted a sensitivity analysis to assess the robustness of our findings to variations in the key parameters. Through these analyses, we aim to provide comprehensive insights into the effectiveness and robustness of the proposed approach for mitigating the impact of disruption events on RER B services. See \newref{Table}{table2}. \\ 
The algorithm was coded using IBM’s CPLEX optimization solver 12.8, utilizing its Python 3.11 API docplex to define the problem’s parameters, objective function, and constraints. Leveraging this robust solver, we efficiently computed the optimal solution, ensuring cost-effective and timely resolution of the problem. with a computer equipped with an Intel Core i7 processor (four cores, 2.50 GHz),  32 GB of RAM, and 1 TB SSD storage on Windows 11 Professional 64-bit.

\subsection{RaaS framework validation}

To validate the efficacy of our RaaS framework, we compared the performance metrics across three key scenarios: normal, do nothing, and RaaS.

\begin{table}[h!]
    \centering
    \caption{Simulation and cost optimization measures for passengers and disrupted operator under normal and disrupted scenarios}
    \begin{tabular}{|l|l|l|l|}
        \hline
        \multicolumn{1}{|c|}{\textbf{Indicator}} & \multicolumn{3}{c|}{\textbf{Scenario}} \\
        \textbf{} & \textbf{Normal} & \textbf{Do-Nothing} & \textbf{RaaS} \\
        \hline
        Average Travel Duration (hh:mm:ss) & 2:02:28 & 3:10:42 & 2:28:46 \\
        Average Wait Duration (hh:mm:ss) & 0:28:46 & 1:16:19 & 0:39:52 \\
        Average Travel Distance (km) & 48.33   & 56.04   & 53.68   \\
        Monetary cost Z1 (euro)         & 650.52  & -       & 1724.1 \\
        Loyalty cost Z2 (euro)          & -       & 6800   & 674.5 \\
        Total cost Z1+Z2 (euro)         & 650.52  & 6800   & 2398.6 \\
        \hline
    \end{tabular}
    \label{table4}
\end{table}

\newref{Table}{table4} presents the simulation and optimization cost measures for passengers and the disrupted operator under normal and disrupted scenarios. In the normal operational scenario, passengers experienced an average travel duration of 2 h, 2 min, and 28 s with a wait duration of 28 min and 46 s, covering an average distance of 48.33 km. However, in a disrupted situation without a replacement process (do-nothing strategy), these metrics increased significantly, with a 55.76\% increase in average travel duration, a 62.82\% increase in waiting duration, and a 16.20\% increase in travel distance. Introducing RaaS intervention in the disrupted scenario demonstrated a significant improvement, reducing the average travel duration by 24.34\%, waiting for the duration by 47.72\%, and maintaining a balanced travel distance compared with the do-nothing scenario. \\

Additionally, the table provides insights into the monetary and loyalty costs incurred by the operator in each scenario, shedding light on the financial implications of the disruption management strategies. In the normal scenario, where operations proceed without any disruptions, the operator incurs a monetary cost of 650.52 euros, reflecting standard operational expenses. Notably, no loyalty costs are incurred in this scenario, indicating minimal passenger dissatisfaction. By contrast, the do-nothing scenario results in significantly heightened loyalty costs, totaling 6800 euros. Furthermore, the RaaS optimization process incurs a monetary cost of 1724.1 euros and notably reduces loyalty costs to 674.5 euros. \\
A comparison of scenarios highlights the effectiveness of the RaaS framework in mitigating the disruption impacts on RER B services. By deploying alternative transportation options, such as bus lines from the nearest stations, the RaaS strategy reduces both travel and wait durations, ensuring passenger satisfaction and loyalty. Moreover, the optimization process ensures cost-effective disruption management, striking a balance between operational efficiency and financial considerations.

\begin{figure}[h!]
\centering
\includegraphics[width=0.74\linewidth]{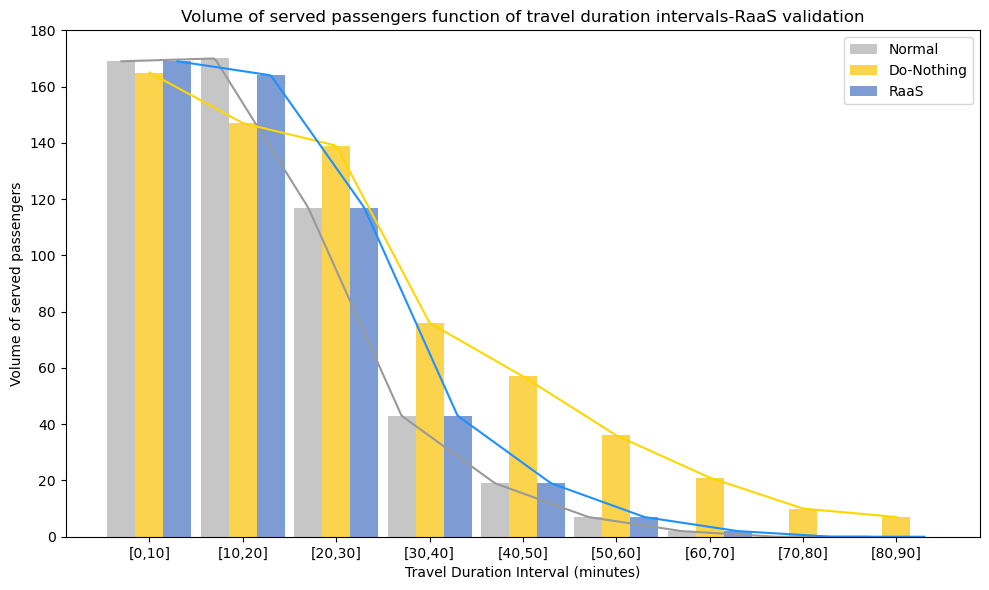}
\caption{Comparative analysis of passenger travel duration distribution}
\label{Figure3}
\end{figure}

\newref{Figure}{Figure3} provides a visual representation of passenger service levels across normal, do-nothing, and RaaS scenarios. When analyzing trip duration distributions across these three operational scenarios, the normal strategy features trips clustered to the left side of the graph, indicating the highest level of service with a significant volume of passengers served in less than one hour. Similarly, the RaaS strategy reflects this trend of shorter trip durations, suggesting that it replicates the efficiency of normal operations, ensuring a swift completion of most trips, even during disruptions. By contrast, the do-nothing strategy exhibited a noticeable number of trips extending beyond one hour, revealing prolonged trip durations without intervention during disruptions. This analysis underscores that the RaaS scenario reduces the disruption impact and maintains the service level by maintaining a higher volume of passengers served across time intervals compared to the do-nothing strategy, which suggests prolonged travel durations without intervention.

\subsection{Benchmarking strategies}

In this section, we consider the existing mitigation scenarios defined in Section 7 and compare them with the RaaS scenario. By benchmarking against established strategies such as bus bridging, taxi bridging, and automated van bridging, we can assess the relative performance of RaaS in terms of passenger service levels and cost efficiency. \newref{Table}{table5} provides a comprehensive comparison of these strategies, highlighting RaaS’s effectiveness in improving travel durations, reducing waiting periods, and optimizing overall costs.

\begin{table}[H]
\centering
\caption{Comparison of RaaS strategy with bus, taxi, and automated van bridging strategies}
\resizebox{\textwidth}{!}{\begin{tabular}{|l|c|c|c|c|}
\hline
\textbf{Indicator} & \multicolumn{4}{c|}{\textbf{Scenario}} \\
%\cline{2-5}
& \textbf{RaaS} & \textbf{Bus Bridging} & \textbf{Taxi Bridging} & \textbf{Automated Van Bridging} \\
\hline
Average Travel Duration (hh:mm:ss) & 2:28:46 & 3:27:39 & 2:29:13 & \textbf{2:22:53} \\
Average Wait Duration (hh:mm:ss) & 0:39:52 & 1:03:15 & \textbf{0:28:16} & 0:30:52 \\
Average Travel Distance (km) & 53.68 & 81.68 & 54.68 & \textbf{53.18} \\
Avg. Arrival Duration of replacement vehicles (hh:mm:ss) & \textbf{00:04:16} & 01:06:00 & 00:04:03 & 00:10:00 \\
Number of reallocated vehicles & 4 & \textbf{3} & 67 & 32 \\
Monetary cost Z1 (euro) & \textbf{1724.1} & 4209.2 & 73431.96 & 2101.25 \\
Loyalty cost Z2 (euro) & \textbf{674.5} & 2018.5 & 796.8 & 988.1 \\
Total cost Z1+Z2 (euro) & \textbf{2398.6} & 6227.7 & 74228.76 & 3089.35 \\
\hline
\end{tabular}}
\label{table5}
\end{table}

In contrast to bus bridging, the RaaS model achieved a 42.14\% reduction in average travel duration and a 37.83\% decrease in wait duration, and the travel distance was 34.65\% shorter with RaaS. It deploys replacement vehicles, taking only 4 min and 16 s on average, which are substantially faster than the 1 h and 6 min required for bus bridging, respectively. RaaS operates with four vehicles, one more than the bus bridge. This discrepancy is attributed to the higher volume of departing passengers in the bus bridging strategy, resulting from the longer wait time. Consequently, the number of passengers remaining at the disrupted station decreases during bus bridging, enabling RaaS to serve numerous stranded passengers. Cost-wise, RaaS is markedly more affordable, with a monetary cost of 59.35\% and a loyalty cost of 74.57\%, compared to bus bridging. \\
By contrast, RaaS marginally improves the average travel duration by 0.20\% and increases the wait duration by 2.66\%. The distance covered was 0.20\% smaller. RaaS slightly outperformed taxi bridging in terms of dispatching vehicles. Remarkably, RaaS requires only four vehicles, rather than the 67 required for taxi bridging. This efficiency translates into a 97.04\% reduction in monetary costs and a 17.41\% decrease in loyalty costs. Our optimizer does not recommend using taxis for replacement in our case study, as its cost is significantly higher than that of the do-nothing strategy. \\
In comparison to automated van bridging, RaaS shows a nominal 0.34\% drop in the average travel duration and a 13.07\% decrease in wait duration. The travel distance decreases by 0.50\%. RaaS’s rapid vehicle dispatch time of 4 min and 16 s considerably outperformed automated van bridging’s 10-minute average. With only four vehicles needed compared to Automated Van Bridging’s 32, RaaS demonstrated superior efficiency in terms of vehicle use. Economically, RaaS was significantly more viable, with a 17.96\% lower monetary cost and a 31.80\% reduction in loyalty costs. \\
In conclusion, our comparative analysis highlights the effectiveness of the RaaS strategy when benchmarked against common bridging strategies, such as bus, taxi, and automated van bridging. RaaS consistently demonstrates superior performance by reducing the average travel and waiting durations while also covering shorter distances. The swift dispatch of replacement vehicles and efficient utilization of resources make RaaS a more economical option, substantially reducing both monetary and loyalty costs.

\begin{figure}[h!]
\centering
\includegraphics[width=0.74\linewidth]{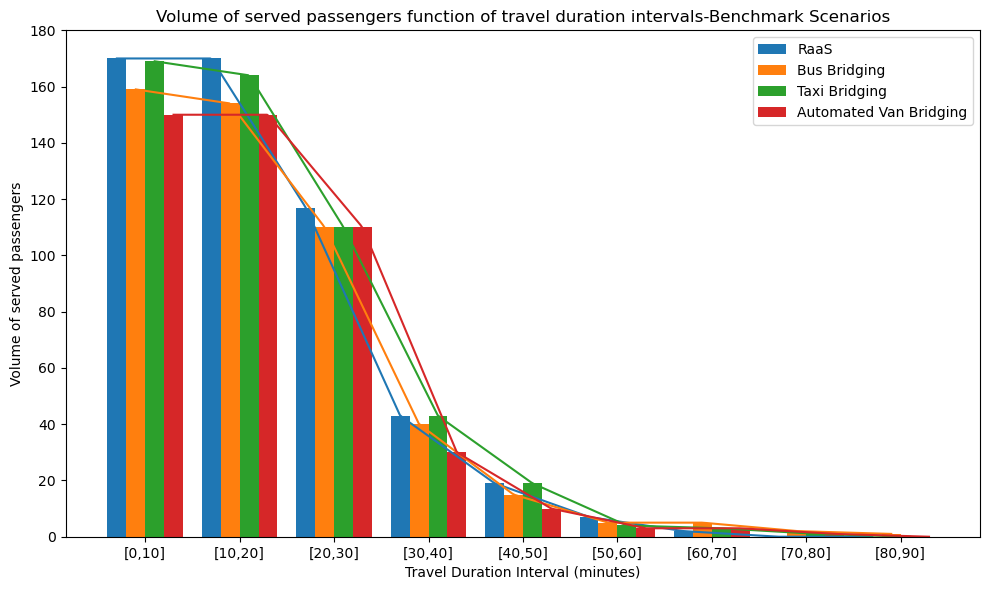}
\caption{Efficiency of transportation strategies through passenger travel duration distribution}
\label{Figure4}
\end{figure}

\newref{Figure}{Figure4} presents a visual comparison of the RaaS with other strategies during service disruptions. Analyzing the efficiency of each transportation strategy based on trip duration revealed distinct patterns. In short, trips lasting 0--30 minutes, RaaS exhibits the highest efficiency by serving the most passengers, followed by taxi bridging and bus bridging, whereas automated van bridging displays the lowest effectiveness. For medium trips spanning 30-60 minutes, RaaS maintains its lead, whereas bus bridging and taxi bridging exhibit declining effectiveness, and automated van bridging continues to lag. For long trips exceeding 60 min, RaaS showed minimal trips surpassing one hour, demonstrating its ability to reduce travel durations. Conversely, automated van bridging and taxi bridging struggle to manage longer durations, whereas bus bridging requires a significant number of long-duration trips, indicating a comparatively lower efficiency. The data depicted in the graph highlight that RaaS consistently served a higher volume of passengers within shorter travel duration intervals, demonstrating its efficiency in minimizing travel and wait durations during disruptions. 

\subsection{Sensitivity analysis}

This section delves into the dynamics of disruption management strategies, examining how varying parameters influence the overall efficiency of different strategies. Through a series of sub-analyses, we explore the effects of different volumes of blocked passengers, the departure rate of passengers during disruptions, and the arrangement rate for replacement services on both monetary and loyalty costs. 

\subsubsection{Effect of the volume of blocked passengers on the monetary and loyalty costs}
In this subsection we evaluate the effectiveness of RaaS in terms of monetary and loyalty costs under the variation of the volume of stranded passengers in the disrupted station. Considering the average demand at the disrupted station $V^n_{ij}$ = 300, for sensitivity analysis, we explore scenarios ranging from 1/3 of this volume as a lower bound to $3*V^n_{ij}$ as an upper bound. Herein, this volume is set from 100 to 900 separated by 200 as an increment, while keeping all the other parameters as in \newref{Table}{table2}. The monetary and loyalty costs are compared under each scenario and the results are shown in \newref{Figure}{Figure5}.

\begin{figure}[!h]
 \centering
 \begin{subfigure}{.5\textwidth}
   \centering
   \includegraphics[width=\linewidth]{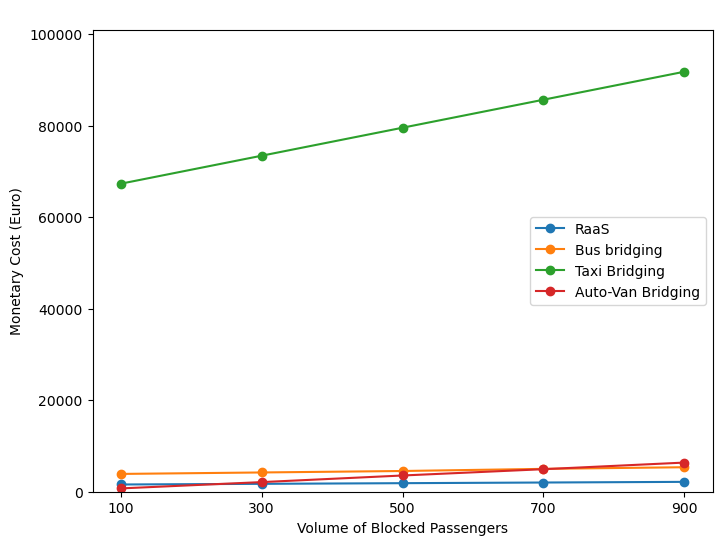}
   \caption{Effect of the volume of blocked passengers on the monetary cost.}
   \label{Fig5a}
 \end{subfigure}%
 \begin{subfigure}{.5\textwidth}
   \centering
   \includegraphics[width=\linewidth]{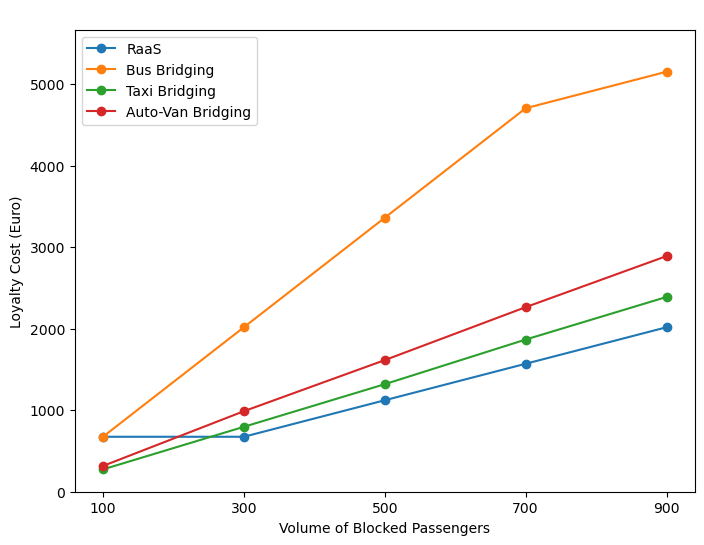}
   \caption{Effect of the volume of blocked passengers on the loyalty cost.}
   \label{Fig5b}
 \end{subfigure}
 \caption{Effect of the volume of blocked passengers on the monetary and loyalty costs.}
 \label{Figure5}
\end{figure}

\newref{Figure}{Fig5a} illustrates that for passenger volumes between 100 to 300, Automated Van Bridging is the most cost-effective strategy, with the lowest cost. However, the cost rises sharply beyond 300 passengers, exceeding those of RaaS and Bus Bridging. RaaS maintains a consistent cost regardless of passenger volume, while Bus Bridging’s costs moderately increase with higher volumes. Taxi Bridging is the least economical, with high costs that escalate as passenger numbers grow. In summary, regarding monetary cost, Automated Van is optimal for up to 300 passengers, beyond which RaaS is the most economical, followed by Bus Bridging. Taxi Bridging is not recommended due to its prohibitive costs.
Meanwhile, \newref{Figure}{Fig5b} shows that for 100 to 200 passengers, Taxi Bridging is the most cost-effective, closely followed by Automated Van Bridging. Beyond 200 passengers, RaaS becomes the most economical due to stable costs, while Taxi and Automated Van Bridging become less favorable. Bus Bridging remains the least beneficial option across all volumes because of its higher costs and time requirements. Therefore, regarding loyalty cost, Taxi Bridging and Automated Van are optimal for smaller volumes of blocked passengers, but RaaS is best for larger volumes, with Bus Bridging being the least favorable.\\
In conclusion, the sensitivity analysis reveals distinct boundaries for the effectiveness of RaaS strategy based on the volume of blocked passengers. In terms of monetary cost RaaS becomes the most cost-effective strategy for volumes exceeding 300 passengers, while in terms of loyalty cost, RaaS becomes most effective for volumes exceeding 200 passengers due to stable costs. This underscores the adaptability of the RaaS strategy, ensuring that it offers a tailored and cost-efficient response to varying scales of disruptions within public transport systems. 

\subsubsection{Effect of the leaving rate on the objective function}

This subsection examines the impact of the leaving rate on the costs incurred during disruption. It aims to identify the optimal leaving rate that minimizes both monetary and loyalty costs, thereby determining the most effective replacement strategy under the variation of this parameter.

\begin{figure}[!h]
 \centering
 \begin{subfigure}{.5\textwidth}
   \centering
   \includegraphics[width=\linewidth]{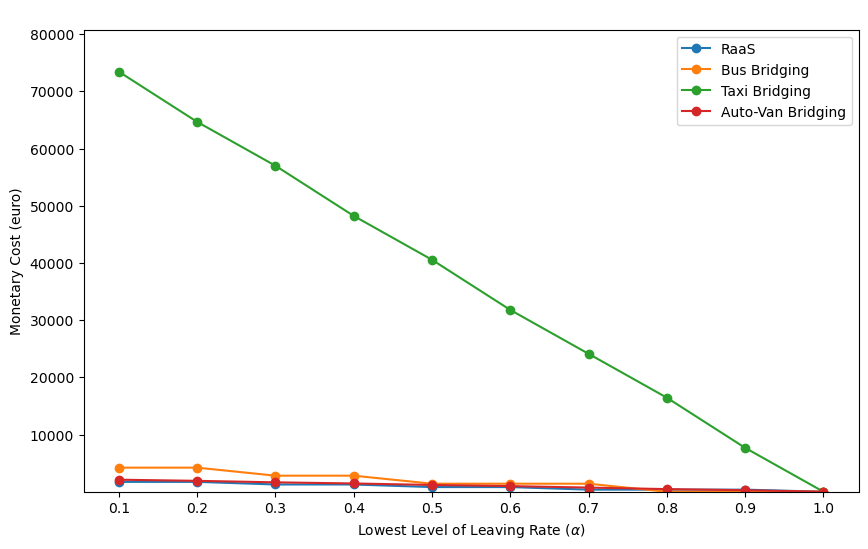}
   \caption{Effect of the lowest level of leaving rate on the monetary cost.}
   \label{Fig6a}
 \end{subfigure}%
 \begin{subfigure}{.5\textwidth}
   \centering
   \includegraphics[width=\linewidth]{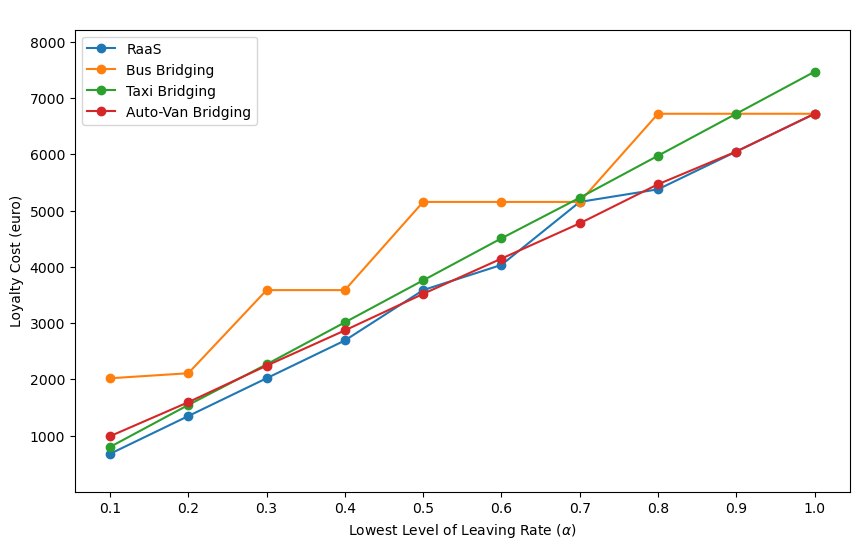}
   \caption{Effect of the lowest level of leaving rate on the loyalty cost.}
   \label{Fig6b}
 \end{subfigure}
  \begin{subfigure}{.5\textwidth}
   \centering
   \includegraphics[width=\linewidth]{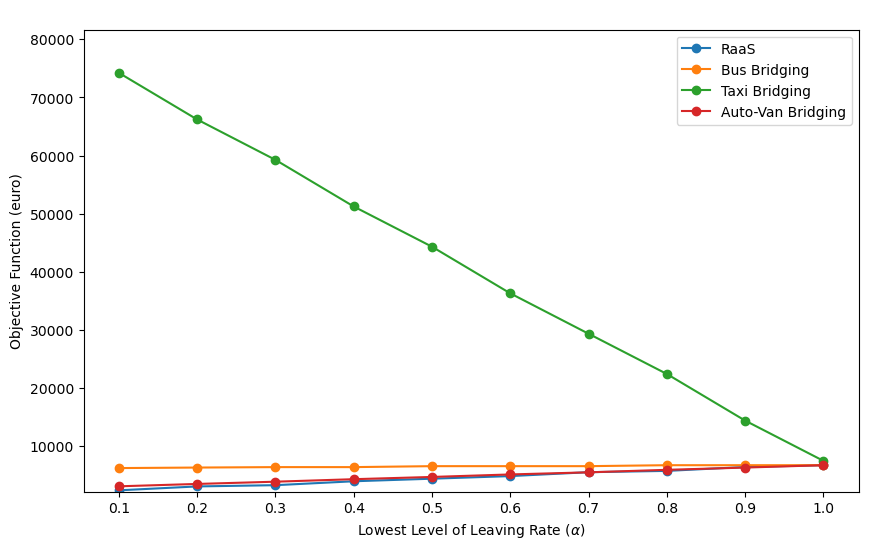}
   \caption{The impact of the lowest level of leaving rate variation on the objective function.}
   \label{Fig6c}
 \end{subfigure}
 \caption{Effect of the lowest level of leaving rate on the monetary and loyalty costs.}
 \label{Figure6}
\end{figure}

The relationship between the lowest level of leaving rate $\alpha$, which determines the departure rate of passengers during disruption $L_{ij}$ and the objective function aiming to minimize the monetary and loyalty costs for different replacement strategies is depicted in \newref{Figure}{Figure6} that includes three sub-figures illustrating the effects of variating $\alpha$ on the monetary cost (\newref{Figure}{Fig6a}), on the loyalty cost (\newref{Figure}{Fig6b}),  and on the objective function (\newref{Figure}{Fig6c}). \\
As the lowest level of leaving rate rises, the operator incurs lower monetary costs due to fewer passengers needing transfer. However, this also leads to increased loyalty costs as more passengers depart, potentially eroding their loyalty to the operator. The objective function seeks a balance between these costs in relation to this parameter.\\
To identify the replacement strategy that can serve the maximum number of blocked passengers with the minimum cost, we first identify the lowest $\alpha$ level for each service where the objective function falls below the cost of Do-Nothing strategy, set at 6800 euros for a volume of blocked passengers of 300 as in \newref{Table}{table4}. Then, we compare the objective function values for each service at this minimum $\alpha$ level to determine the best scenario. Across all services, including RaaS, Bus Bridging, Taxi Bridging, and Automated Van Bridging, the objective function remains below 6800 euros from $\alpha = 0.1$ onwards, making this the optimal $\alpha$ level for each service. Comparing the objective function values at $\alpha = 0.1$, RaaS demonstrates the lowest cost with 2398.6 euros, followed by Automated Van Bridging with 3089.35 euros, and Bus Bridging comes in the third place with 6227.74 euros. Since Taxi Bridging's objective function far exceeds the threshold, it is excluded from consideration. Thus, RaaS at $\alpha = 0.1$ emerges as the best scenario, meeting the criteria of having the minimum volume of leaving passengers and the minimum objective cost. 

\subsection{Effect of the arrangement rate on the objective function}

In assessing the cost-effectiveness of various replacement strategies, we examine their performance relative to the Do-Nothing strategy across different rates $CA^{m,x}$ of calculating the arrangement cost $CA(\gamma^{rs,m,x}_{ij})$ and volumes of stranded passengers $V^n_{ij}$. If the objective function’s value surpasses the Do-Nothing cost, the strategy is deemed not cost-effective, because this indicates that the operator would face greater losses compared to opting out of implementing a replacement service during the disruption. Employing this approach provides an overview of each strategy’s ability to alleviate the effects of disruptions, factoring in the costs involved.\\ 
\newref{Figure}{Figure7} illustrates how the arrangement rate influences the cost-effectiveness of various replacement strategies during transportation disruptions. Each sub-figure corresponds to a different volume of stranded passengers, ranging from 100 to 900 (sub-figures (\textit{a}) to (\textit{e})) for arrangement rates ranging from 0.1 (10\%) to 1 (100\%), and compares the objective function values to the Do-Nothing strategy cost.

\begin{figure}[!h]
 \centering
 \begin{subfigure}{.5\textwidth}
   \centering
   \includegraphics[width=\linewidth]{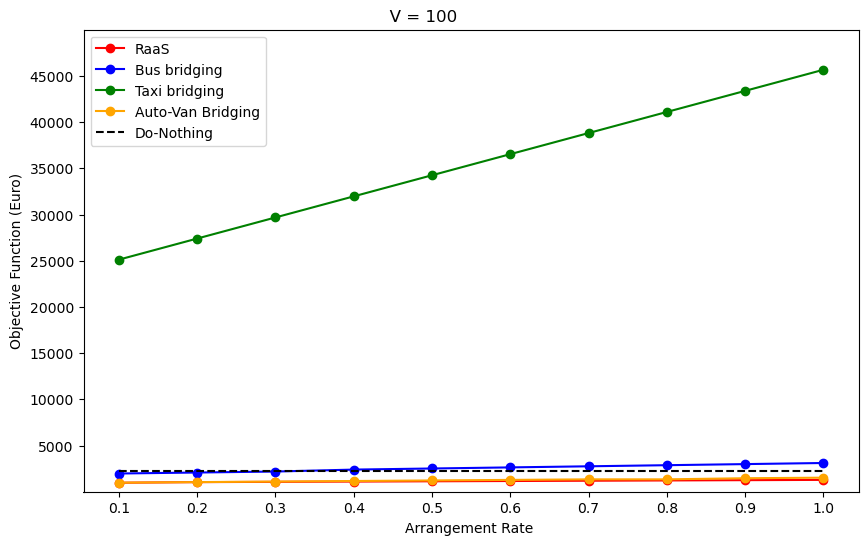}
   \caption{}
   \label{Fig7a}
 \end{subfigure}%
 \begin{subfigure}{.5\textwidth}
   \centering
   \includegraphics[width=\linewidth]{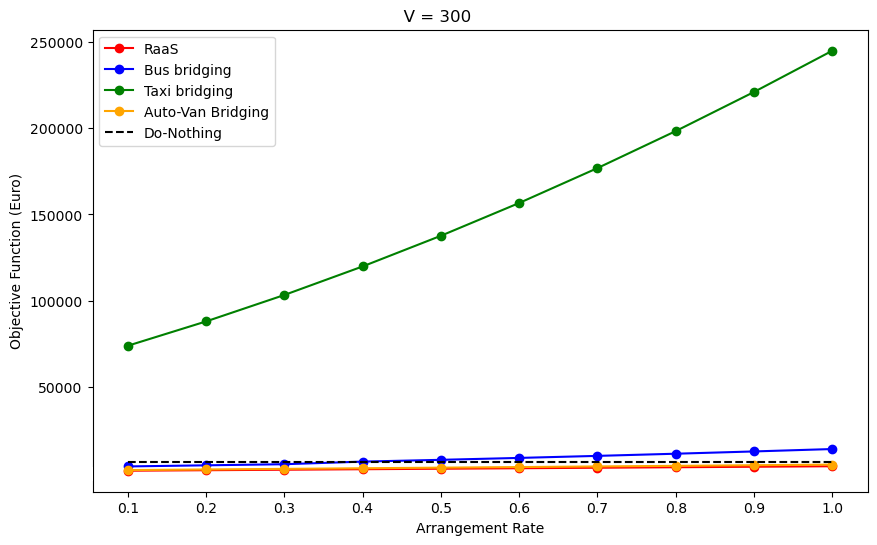}
   \caption{}
   \label{Fig7b}
 \end{subfigure}

 \begin{subfigure}{.5\textwidth}
   \centering
   \includegraphics[width=\linewidth]{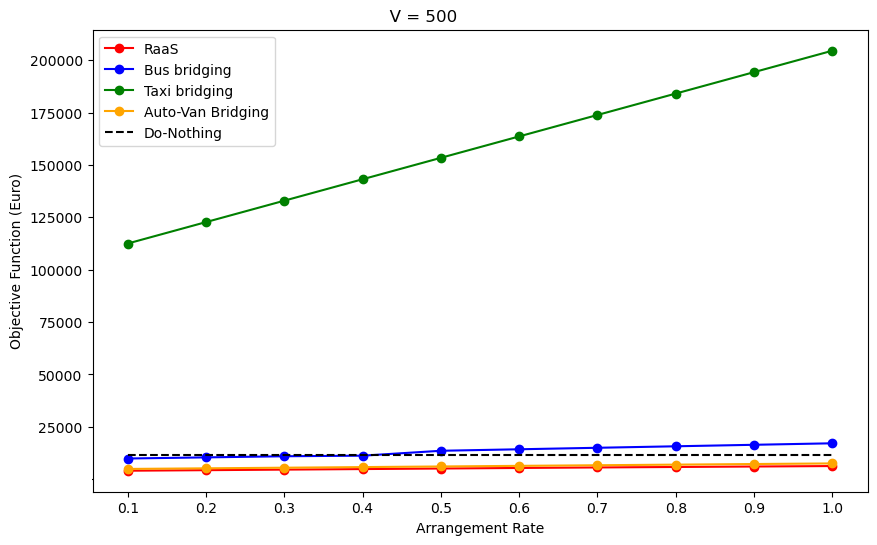}
   \caption{}
   \label{Fig7c}
 \end{subfigure}%
 \begin{subfigure}{.5\textwidth}
   \centering
   \includegraphics[width=\linewidth]{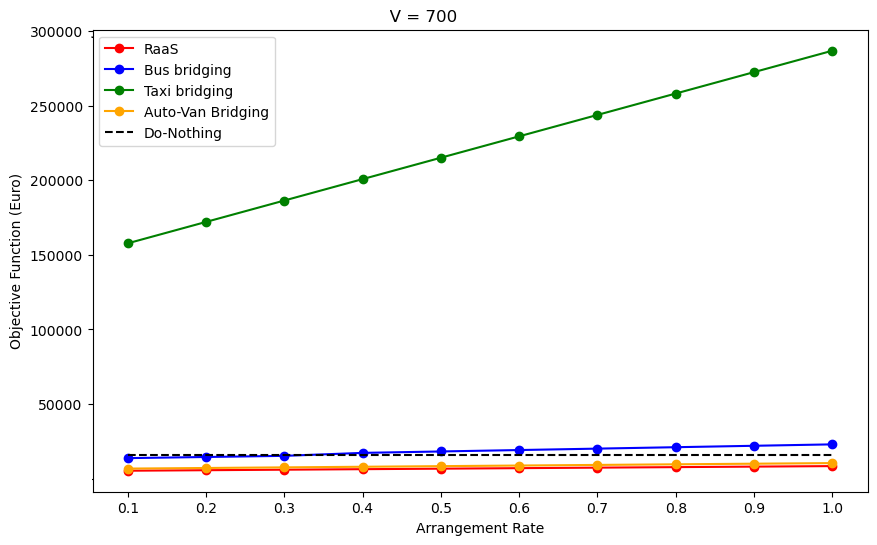}
   \caption{}
   \label{Fig7d}
 \end{subfigure}

 \begin{subfigure}{.5\textwidth}
   \centering
   \includegraphics[width=\linewidth]{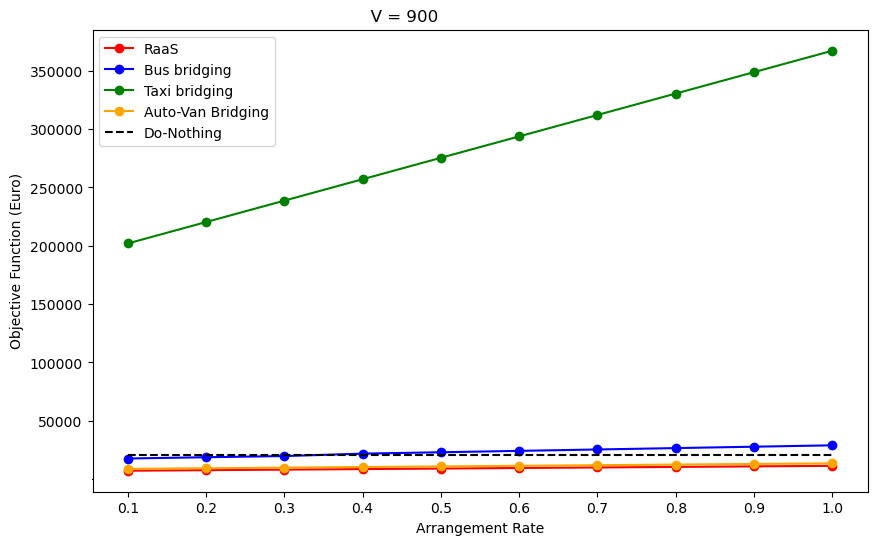}
   \caption{}
   \label{Fig7e}
 \end{subfigure}
 \caption{Effect of the variation of the arrangement rate under each volume of blocked passengers on the objective function.}
 \label{Figure7}
\end{figure}

In the comparative analysis of replacement strategies, spanning various arrangement rates (ranging from 10\% to 100\% of each strategy's service cost) and different volumes of stranded passengers (100, 300, 500, 700, and 900), as illustrated in Figure 7, the objective function values depict the total cost, inclusive of monetary and loyalty costs, associated with implementing each strategy relative to the Do-Nothing strategy. 
For V=100, RaaS emerges as the most cost-effective strategy, consistently yielding objective function values below the Do-Nothing cost of 2265 euros across all arrangement rates; for instance, at an arrangement rate of 0.1, RaaS incurs a cost of 972.14 euros. Automated Van Bridging closely follows suit in cost-effectiveness, with a slight increase in cost, starting at 979 euros at an arrangement rate of 0.1. Conversely, Bus Bridging initiates with a higher cost of 1970.8 euros at an arrangement rate of 0.1, progressively escalating and surpassing the Do-Nothing cost at an arrangement rate of 0.4. Taxi Bridging, on the other hand, emerges as the least cost-effective, with costs starting at 25116.652 euros at an arrangement rate of 0.1, significantly exceeding the Do-Nothing cost. 
For V=300, RaaS sustains its lead in cost-effectiveness, with costs beginning at 1724.07 euros at an arrangement rate of 0.1, while Automated Van Bridging remains a feasible option, with costs starting at 2101.24 euros at an arrangement rate of 0.1. However, Bus Bridging demonstrates diminishing cost-effectiveness with increasing volume, with costs beginning at 4209.2 euros at an arrangement rate of 0.1, and Taxi Bridging continues to exhibit inefficiency, with costs soaring to 74043.8901 euros at an arrangement rate of 0.1. 
For V=500, RaaS persists in offering the lowest costs, starting at 4016.3 euros at an arrangement rate of 0.1, while Automated Van Bridging experiences a slight increase in costs, beginning at 4866 euros at an arrangement rate of 0.1. Both Bus Bridging and Taxi Bridging considerably surpass the Do-Nothing loss, with Taxi Bridging reaching an exorbitant cost of 112522.601 euros at an arrangement rate of 0.1. 
For V=700 and V=900, the observed trends reflect those of lower volumes, with RaaS and Automated Van Bridging maintaining their status as the most cost-effective strategies, whereas Bus Bridging rapidly loses its cost-effectiveness with escalating arrangement rate, and Taxi Bridging remains the least efficient, with costs increasing significantly. \\
In summary, RaaS consistently emerges as the most cost-effective strategy across all volumes and arrangement rates, closely followed by Automated Van Bridging, while Bus Bridging may warrant consideration at lower volumes and arrangement rates, and Taxi Bridging should be avoided due to its disproportionate costs relative to the benefits offered. The results underscore that as the volume of stranded passengers and the arrangement rate escalate, the cost-effectiveness of RaaS and Automated Van Bridging become even more pronounced, rendering them the preferred strategies for managing transportation disruptions under the variation of these two parameters.

\section{Discussion and Conclusion}

To enhance the resilience of public transport systems, this study challenged the dual objectives of optimizing resource utilization and ensuring uninterrupted services and passenger satisfaction during unpredictable events. These demands vary, requiring a model that can reconcile the diverse interests of stakeholders: minimizing financial costs for operators, while minimizing time costs for passengers.
To address these challenges, we developed RaaS framework designed to be adaptable, scalable, and efficient for managing public transport disruptions. By integrating various transportation modes and utilizing existing resources more effectively, RaaS enhances the overall resilience of public transport systems during unexpected challenges. Our model is validated and benchmarked against conventional strategies, and sensitivity analyses are conducted to assess its effectiveness under various parameter conditions. 

The validation of the RaaS framework (\newref{Table}{table4}) reveals its significant efficacy in reducing the average travel duration by 24.34\% and wait duration by 47.72\% while maintaining a balanced travel distance when compared to the do-nothing scenario. This validation was exemplified by the swift deployment of four bus lines from the nearest stations, resulting in an average vehicle arrival duration of approximately 4 min, showcasing RaaS’s prompt response to disruptions. When benchmarked against common strategies (\newref{Table}{table5}), RaaS demonstrated a substantial reduction in both the average travel duration (42.14\%) and wait duration (37.83\%) compared with Bus Bridging. Furthermore, it outperformed taxi bridging with significant reductions in both monetary (97.04\%) and loyalty (17.41\%) costs. By contrast, RaaS showed a nominal 0.34\% drop in average travel duration and a 13.07\% decrease in wait duration, with a 17.96\% lower monetary cost and a 31.80\% reduction in loyalty costs. 

Sensitivity analysis further confirmed RaaS’s position as the most cost-effective strategy. This shows the consistent ability to scale effectively to different passenger volumes (\newref{Figure}{Figure5}), particularly in scenarios with larger groups of commuters. The RaaS framework also proved its merit in fostering passenger loyalty, particularly at lower rates of passenger departure during disruptions (see \newref{Figure}{Figure6}). This indicates the potential of RaaS’s potential to maintain high levels of passenger retention and satisfaction even in challenging circumstances. Moreover, the analysis of arrangement rates and their impact on the overall cost-effectiveness of the various strategies highlights RaaS’s financial viability (\newref{Figure}{Figure7}). It remains an optimal choice across a spectrum of arrangement rates and passenger volumes, in contrast to other strategies that may offer only limited cost benefits under specific conditions.  %In summary, a comprehensive examination of the RaaS framework and its comparison with alternative strategies such as bus bridging, taxi bridging, and automated van bridging reaffirms its robustness and superiority in managing disruptions within transportation systems. From substantial reductions in the average travel and wait durations to consistent cost-effectiveness across varying passenger volumes and arrangement rates, RaaS has emerged as a resilient and financially viable solution. 

The success of the RaaS framework underscores the need for ongoing research and development in this domain, with a focus on real-time monitoring, predictive analytics, and adaptive response mechanisms. Integration of automated vehicles presents a transformative opportunity, offering dynamic solutions to disruptions through vehicles that can swiftly adapt to changing circumstances and potentially provide more affordable solutions to further enhance the resilience of transportation systems.

%%%%%%%%%%%%%%%%%%%%%%%%%%%%%%%%%%%%%%%%%%%%%%%%%%%%%%%%%%%%%
%% BIBLIOGRAPHY
%%%%%%%%%%%%%%%%%%%%%%%%%%%%%%%%%%%%%%%%%%%%%%%%%%%%%%%%%%%%%

%\bibliographystyle{elsart-harv}
%\setlength{\bibsep}{0pt}
% \newpage
\bibliography{biblio1}

% End line numbering
%\nolinenumbers

\end{document}